\documentclass[10pt,twocolumn,letterpaper]{article}

\usepackage{cvpr} % uncomment this for the final submission
\usepackage{times}
\usepackage{epsfig}
\usepackage{graphicx}
\usepackage{amsmath}
\usepackage{amssymb}

\usepackage[table]{xcolor} % Allows cell coloring
\usepackage{colortbl}      % Additional table coloring support
\usepackage{booktabs}      % For clean table formatting
\usepackage{makecell}
\usepackage{multirow}
\usepackage{tabularx}
\usepackage{float}
\usepackage{pifont}
\usepackage{graphbox}

\usepackage[pagebackref=true,breaklinks=true,letterpaper=true,colorlinks,bookmarks=false]{hyperref}

\definecolor{header}{gray}{0.90}

\begin{document}

%%%%%%%%% TITLE
\title{What's Old is New Again: Classical Dimensionality Reduction \\ for Efficient Saliency-Guided Biometric Attack Detection}

\author{Samuel Webster\hspace{2cm}Walter Scheirer\\
Department of Computer Science and Engineering\\
University of Notre Dame, IN, USA\\
{\tt\small \{swebster,wscheire\}@nd.edu}
}

\maketitle
\thispagestyle{empty}

%%%%% ABSTRACT
\begin{abstract}
    Saliency-guided training is a paradigm in visual recognition that encourages models to focus on the most relevant image regions during learning. While its application in biometric presentation attack detection (PAD) has shown strong benefits in robustness and generalization, adoption is often limited by the high cost, domain specificity, and limited scalability of existing saliency acquisition methods, such as human annotations over a limited dataset. We present a novel, cost-efficient, and highly-scalable approach to saliency acquisition using maps inspired by classical dimensionality reduction techniques: PCA and LDA. Our proposed methods generate saliency maps directly from raw training data, requiring no human annotation nor domain knowledge. We contextualize the effectiveness of these saliency sources in three saliency-explored domains (iris PAD, synthetic face detection, fingerprint PAD) and demonstrate its scalability in two saliency-novel domains (fingerprint vein PAD and ID card PAD). Across all domains tested, models trained using dimensionality reduction-sourced saliency maps exceed baseline and sometimes SOTA saliency methods without any resource investment or domain-specific tooling. Our findings overcome an important yet unaddressed barrier to saliency-guided training for biometric attack detection and beyond.
\end{abstract}

%%%%% BODY TEXT

% Introduction
\section{Introduction}

Saliency-guided training has been successfully implemented in multiple biometric presentation attack detection (PAD) domains \cite{boyd2022human, boyd2023cyborg, webster2025saliency}. By instructing convolutional neural networks ``where to look'' during training, saliency guidance facilitates improvements to decision explainability as well as robustness and generalization to unseen attack types. Saliency-guided training is highly dependent on configuration, such as fidelity of saliency maps \cite{crum2024grains} or loss-based \cite{boyd2023cyborg} versus transformation-based\cite{boyd2022human} region guidance.

\begin{figure}
\centering

\includegraphics[width=0.45\textwidth]{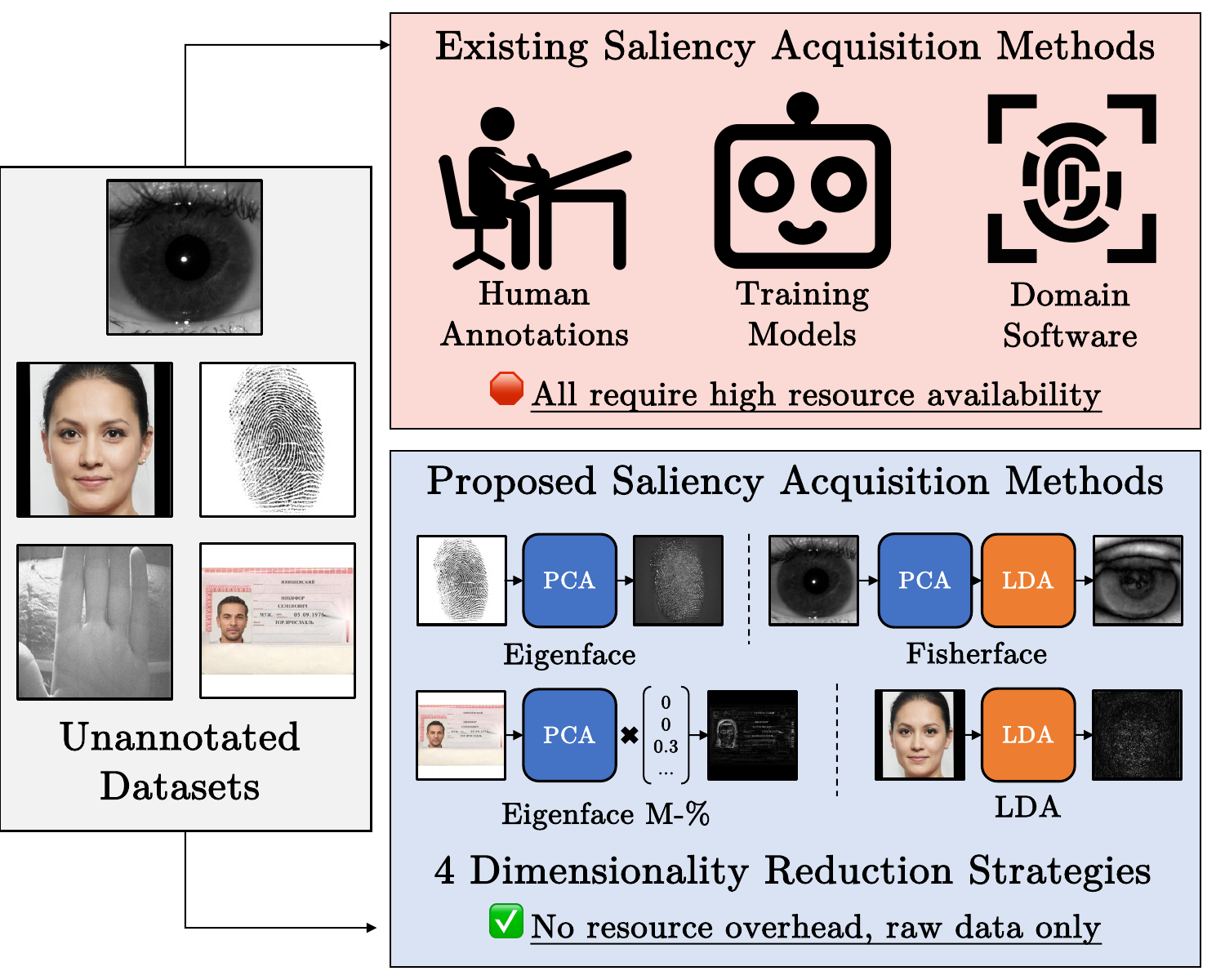}
\caption{\textbf{A brief comparison between existing saliency strategies and our proposed method.} While existing pipelines contain high-expense human annotation collection studies or specific domain technologies, our methods building upon classical dimensionality reduction have virtually no barrier to application in any domain.}
\label{fig:teaser}

\end{figure}

However, acquisition of saliency maps are a major bottleneck for applying saliency-guided methods to new datasets or domains. Implementations of saliency-guided training in new domains regularly apply human-annotated saliency maps --- pixel-wise annotations of regions deemed important by a human annotator for the given classification task. While these maps support meaningful improvements to model performance and explainability, they are time-consuming, labor-intensive, and expensive to acquire as well as only applicable to the specific set of annotated biometric samples, limiting their scalability and general applicability. Recent research has sought to substitute human annotations with generative saliency maps, such as those produced by autoencoder \cite{crum2024grains}, teacher-student \cite{crum2023teaching}, or domain-specific segmentation \cite{crum2024grains} techniques. Although these approaches reduce annotation costs when applied at scale, they still rely on either a base set of human labels, applicable models and algorithms, or otherwise domain-specific knowledge.

This work proposes an alternative path: per-sample maps inspired from classical dimensionality reduction techniques in computer vision --- Eigenfaces \cite{eigenfaces} and Fisherfaces \cite{fisherfaces} --- as a domain-agnostic, human annotation-free source of saliency. Originally developed for facial recognition, these methods decompose images into meaningful components based on statistical variation, found with Principal Component Analysis (PCA), and discriminative power, found with Linear Discriminant Analysis (LDA). Unlike human annotations or pretrained segmentation models, Eigenface- and Fisherface-derived saliency maps can be generated directly from a dataset without prior human annotation or domain expertise, as outlined in Fig. \ref{fig:teaser}.

We examine four distinct dimensionality reduction-sourced saliency strategies as alternatives to existing saliency sources in guiding deep learning models. We compare their performance against existing saliency acquisition methods by evaluating their application to three saliency-explored biometric domains: iris PAD \cite{boyd2022human, crum2024grains}, synthetic face detection \cite{boyd2023cyborg, crum2024grains}, and fingerprint PAD \cite{webster2025saliency, webster2026psychophysical}. We further validate their scalability to new datasets through their application to two saliency-novel domains: fingerprint vein PAD \cite{candyfv} and identity document PAD \cite{id_document_pad, id_document_pad_2}. If effective, our proposed approach would significantly lower the barrier to entry for saliency-guided training in new domains, promoting accessibility to robust biometric models. 

\subsection{Research Questions}

We apply our proposed saliency-acquisition methods based in classical dimensionality reduction techniques and answer the following research questions:

\begin{enumerate}
    \item \textbf{In biometric attack detection domains where saliency-guided training has been previously applied, do our proposed dimensionality reduction-sourced saliency maps guide models to competitive performance?}
    \item \textbf{Is our proposed dimensionality reduction-based saliency acquisition strategy effective in biometric attack detection domains where saliency-guided training has never been applied?}
\end{enumerate}

We release all relevant source material at \texttt{https://github.com/placeholder}\footnote{Link is redacted for author anonymity. If accepted, we will add the correct GitHub link to the camera-ready version of the paper}, including code for producing all described dimensionality reduction-based saliency maps and training saliency-guided models.

% Related Work
\section{Related Work}

\subsection{Saliency-guided Training for Biometric Attack Detection}

Saliency-guided training incorporates external annotations into the optimization process of convolutional neural networks. In biometric attack detection, per-sample saliency maps aim to identify discriminative areas or features. By synchronizing model decision-making to the provided regions of interest, prior work has reported improvements in both model explainability and robustness to unseen attack instruments \cite{crum2024grains, webster2025saliency}. There are two prevailing manners of implementation: loss-based guidance, which incorporates saliency directly in the training objective \cite{boyd2023cyborg}, and transformation-based guidance, which modifies the input image according to its associated saliency map \cite{boyd2022human}. Comparative implementations suggest that loss-based guidance is more effective because it aligns models while preserving input image fidelity, unlike the lossy nature of transformation approaches \cite{webster2025saliency}.

% Saliency-guided training is a convolutional neural network training strategy that aligns decision-making with annotated regions of interest, as indicated by sample-specific saliency maps. Saliency guidance has demonstrated improvements to explainability as well as generalization, making it suitable for biometric PAD tasks which frequently test against unknown attack instruments \cite{crum2024grains, webster2025saliency}. It is regularly implemented in two manners: loss-based guidance \cite{boyd2023cyborg} and transformation-based \cite{boyd2022human} guidance. Comparative implementations suggest that loss-based guidance is more effective due to the lossy nature of saliency-based image transformation, such as blurring non-salient regions \cite{webster2025saliency}.

CYBORG loss \cite{boyd2023cyborg} is one of the most widely adopted loss-based saliency guidance formulations. It combines two training objectives: a conventional classification objective ($\mathcal{L}_\text{classification}$) and a saliency-alignment objective ($\mathcal{L}_\text{saliency-guidance}$). This approach allows the network to jointly optimize predictive performance and desired region attribution. The resulting loss is defined as:

% This paper applies CYBORG loss \cite{boyd2023cyborg} as a representative loss for saliency-guided training. It is a two-component loss composed of classification component $\mathcal{L}_\text{classification}$ and human saliency component $\mathcal{L}_\text{saliency-guidance}$, formulated as:

\begin{equation}
\begin{split}
\mathcal{L}_\text{CYBORG} = &~\alpha*\mathcal{L}_\text{classification} \\
& + (1-\alpha)*\mathcal{L}_\text{saliency-guidance},\\
\end{split}
\end{equation}

\noindent where the parameter $\alpha$ controls the relative influence of both loss components. Larger values of $\alpha$ prioritize classification performance, whereas smaller values emphasize attributing decisions to annotated regions of interest. In CYBORG, the Class Activation Mapping is fitted to external saliency maps, promoting feature learning in areas identified as informative.

% \noindent where $\alpha$ weighs the two loss components. When $\alpha=1.0$, $\mathcal{L}_\text{CYBORG}$ becomes a standard cross-entropy loss, and when $\alpha=0.0$, $\mathcal{L}_\text{CYBORG}$ is only penalizing the model for lack of alignment between the model's and external saliencies, regardless of classification performance. In CYBORG~\cite{boyd2023cyborg}, saliency-guidance loss aligns Class Activation Mapping with human-sourced saliency provided for all training examples. By balancing human alignment with classification loss, models are more likely to learn discriminative features from human-salient regions. 

In transformation-based saliency guidance, perceptual knowledge is provided directly in the input representation. These methods use saliency maps to alter input images, preserving annotated regions while degrading non-salient regions though operations such as blurring or noise injection  \cite{boyd2022human}. By effectively removing features present in unannotated areas, transformation-based guidance encourages learning only on salient, annotated regions.

Across biometric attack detection domains, saliency-guided training has been explored using a diverse set of saliency sources. Human-drawn annotations have been successfully employed for iris PAD, synthetic face detection, and fingerprint PAD \cite{boyd2022human, boyd2023cyborg, webster2025saliency}. Alternative sources include teacher-student generated saliency maps \cite{crum2023teaching}, domain segmentation model-sourced saliency \cite{crum2024grains}, software-based fingerprint minutiae and quality-map based pseudosaliency \cite{webster2025saliency}, and autoencoder-generated saliency trained to approximate human annotations \cite{crum2024grains, webster2025saliency}. Prior work has additionally examined how the granularity of saliency information influences performance, treating expressivity of annotation strength as a controllable saliency parameter \cite{crum2024grains}.

% Saliency-guided training has demonstrated significant improvements in model generalization and classification accuracy when guided by both human- and software-sourced saliency. Iris PAD models have benefited with guidance by human annotations \cite{boyd2022human}, AI student-generated saliency \cite{crum2023teaching}, and iris segmentation saliency \cite{crum2024grains}. For synthetic face detection, a biometric PAD task, the currently explored saliency types are human annotations \cite{boyd2023cyborg}, AI student-generated saliency \cite{crum2023teaching}, and face segmentation saliency \cite{crum2024grains}. In fingerprint PAD, successful saliency types include human annotations, minutiae-based saliency, and quality-based saliency \cite{webster2025saliency}. For all types, autoencoder-generated saliency has been explored, trained to predict human annotations on new samples \cite{crum2024grains, webster2025saliency}. To maximize performance using saliency-guided training, the fidelity of conveyed information can be optimized through varying saliency granularities, such as: Features of Interest (FOI) provide high-fidelity (0–255) maps, Areas of Interest (AOI) represent binarized maps (0 or 255), and Boundaries of Interest (BOI) define the minimally enclosing rectangle around a saliency map's marked regions \cite{crum2024grains}.

\begin{table*}[t]
\centering
\caption{\textbf{A summary of capabilities and barriers across saliency map acquisition strategies.} Unlike prior works that rely on external bottlenecks (human labor or domain-specific tools), our dimensionality reduction-based approach derives saliency from intrinsic dataset variance, decreasing barriers to applying saliency guidance.}
\label{tab:saliency-methods}

\newcolumntype{L}{>{\raggedright\arraybackslash}X}
\newcolumntype{C}{>{\centering\arraybackslash}X}

\small
\begin{tabularx}{\linewidth}{L L c c L L c}
\toprule
\rowcolor{header}
\textbf{Saliency Strategy} & \textbf{Annotation Content} & \textbf{Resolution} & \textbf{Scalability} & \textbf{Domain Prerequisites} & \textbf{Primary Limitation} & \textbf{Performance} \\
\midrule
Human Annotations \cite{boyd2022human, boyd2023cyborg, webster2025saliency} & Manually Identified Regions & Full & Poor & Data Only & High cost; non-scalable & Above Baseline \\
\addlinespace
Human-mimicking AE \cite{crum2024grains, webster2025saliency} & Predicted Human Annotations & Full & High & Human-annotative Training Data & Reliance on human annotations & Above Baseline \\
\addlinespace
Teacher-Student \cite{crum2023teaching} & Human-like Distillation & Limited & High & Human-annotative Training Data & Initial training samples; resolution loss & Above Baseline \\
\addlinespace
Domain Segmenters \cite{crum2024grains} & Coarse Binary Region & Full & High & Pre-trained Domain Model & Low semantic interpretability & Below Baseline \\
\addlinespace
Domain Software \cite{webster2025saliency} & Domain-specific Regions & Tool-dependent & High & Existing Domain Software & Limited to specific tools & Above Baseline \\
\addlinespace
\midrule
\rowcolor[HTML]{E6F2FF} \textbf{Dim. Reduction (Ours)} & \textbf{Dataset Variance} & \textbf{Full} & \textbf{High} & \textbf{Data Only} & \textbf{Variability represented} & \textbf{Above Baseline} \\
\bottomrule
\end{tabularx}
\end{table*}

Notably, all explored saliency types rely on some level of resource investment or domain knowledge, as summarized in Tab. \ref{tab:saliency-methods}. For human-annotative saliency, produced maps are applicable only in the applied domain \cite{boyd2022human, boyd2023cyborg, webster2025saliency}. Further, human annotation collection is an expensive process, requiring many human annotators and producing limited-sized datasets by modern deep learning standards \cite{crum2024grains}. Software or model-sourced saliency types are an effective scalable alternative \cite{webster2025saliency}. However, these methods still fall short due to their reliance on existing domain tooling or knowledge: iris and face segmenter models must already exist \cite{crum2024grains}, human saliency must exist to train an autoencoder \cite{crum2024grains, webster2025saliency} or apply a teacher-student paradigm \cite{crum2023teaching}, and domain-oriented software must exist to produce pseudosaliency \cite{webster2025saliency}. These high resource barriers make the acquisition of saliency in new domains very difficult when no preexisting annotative material or generative means exist.

\subsection {Eigenfaces and Fisherfaces}

Eigenface and Fisherface techniques are classical methods for face recognition that employ dimensionality reduction to capture salient features of facial images. Eigenfaces, introduced by Turk and Pentland~\cite{eigenfaces}, uses Principal Component Analysis (PCA) to identify the principal axes of variation in a set of face images. By projecting images onto these axes, the method captures the most expressive components of facial appearance, effectively encoding structural and identity-relevant features. The resulting PCA components represent a basis set from which any face in the dataset can be approximately reconstructed.

Fisherfaces, proposed by Belhumeur et al.~\cite{fisherfaces}, builds upon Eigenfaces to enhance class separability by applying Linear Discriminant Analysis (LDA) to the projected faces. Unlike PCA, which simply maximizes image set variance without regard to identity, LDA finds a linear projection that maximizes between-class variance while minimizing within-class variance, emphasizing identity-defining features. When applied to face images, Fisherfaces emphasize features that are most useful for distinguishing between individuals, making them particularly effective for recognition under varying lighting and expression conditions.

While originally developed for classification and identification tasks, both techniques inherently emphasize information-rich regions of the image. Further, they both support reverse transformation processes that produce a mask of composed Eigenfaces and Fisherfaces.

% Methods
\section{Methods}

\begin{figure*}[t]
\centering
\small

\setlength{\tabcolsep}{2pt}

\scalebox{0.80}{
\begin{tabular}{c c c c c c}

\toprule
\rowcolor{header}
\textbf{\large Domain} & \textbf{\large Sample Input} & \textbf{\large Eigenface} & \textbf{\large Eigenface M-0.5} & \textbf{\large Fisherface} & \textbf{\large LDA} \\
\bottomrule
\addlinespace[4pt]

\makecell[c]{\large\bfseries Iris \\ \large\bfseries PAD} &
\includegraphics[width=3cm,align=c]{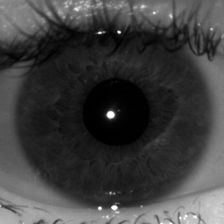} &
\includegraphics[width=3cm,align=c]{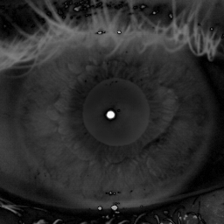} &
\includegraphics[width=3cm,align=c]{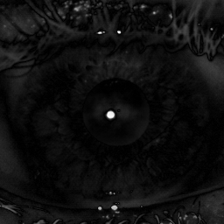} &
\includegraphics[width=3cm,align=c]{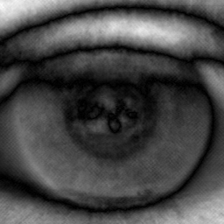} &
\includegraphics[width=3cm,align=c]{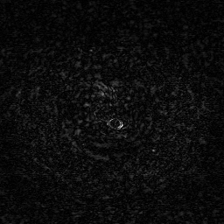} \\
\addlinespace[8pt]

\makecell[c]{\large\bfseries Synthetic \\ \large\bfseries Face \\ \large\bfseries Detection} &
\includegraphics[width=3cm,align=c]{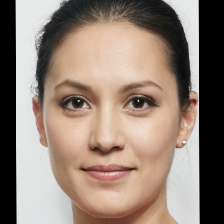} &
\includegraphics[width=3cm,align=c]{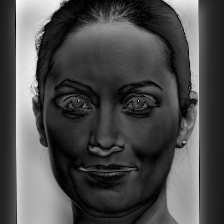} &
\includegraphics[width=3cm,align=c]{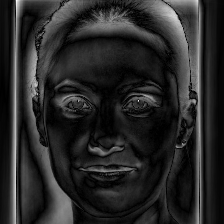} &
\includegraphics[width=3cm,align=c]{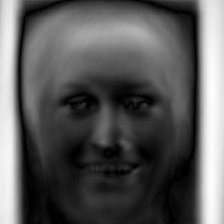} &
\includegraphics[width=3cm,align=c]{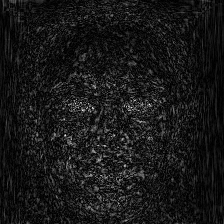} \\
\addlinespace[8pt]

\makecell[c]{\large\bfseries Fingerprint \\ \large\bfseries PAD} &
\includegraphics[width=3cm,align=c]{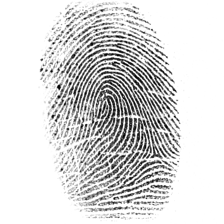} &
\includegraphics[width=3cm,align=c]{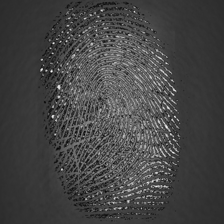} &
\includegraphics[width=3cm,align=c]{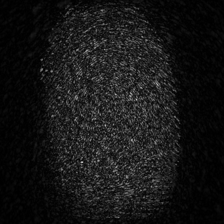} &
\includegraphics[width=3cm,align=c]{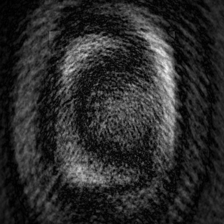} &
\includegraphics[width=3cm,align=c]{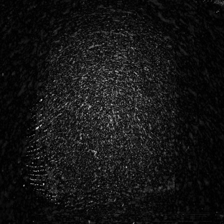} \\
\addlinespace[8pt]

\makecell[c]{\large\bfseries Fingerprint \\ \large\bfseries Vein \\ \large\bfseries PAD} &
\includegraphics[width=3cm,align=c]{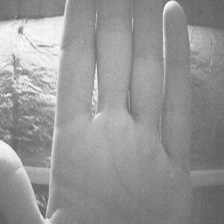} &
\includegraphics[width=3cm,align=c]{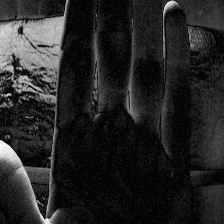} &
\includegraphics[width=3cm,align=c]{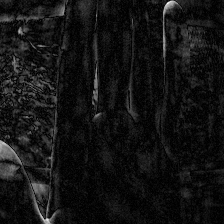} &
\includegraphics[width=3cm,align=c]{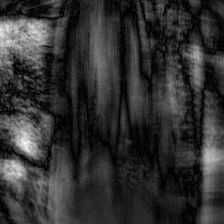} &
\includegraphics[width=3cm,align=c]{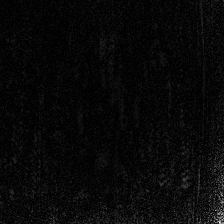} \\
\addlinespace[8pt]

\makecell[c]{\large\bfseries Identification \\ \large\bfseries Document \\ \large\bfseries PAD} &
\includegraphics[width=3cm,align=c]{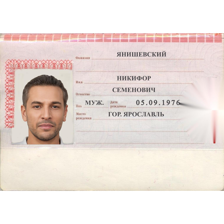} &
\includegraphics[width=3cm,align=c]{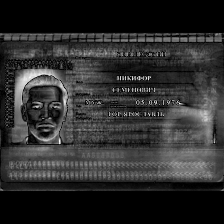} &
\includegraphics[width=3cm,align=c]{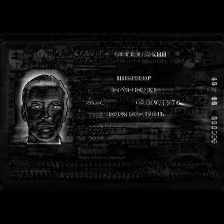} &
\includegraphics[width=3cm,align=c]{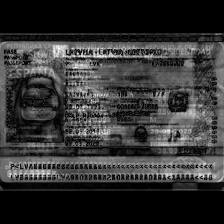} &
\includegraphics[width=3cm,align=c]{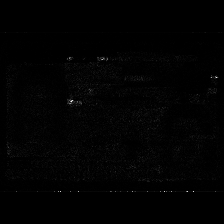} \\
\addlinespace[8pt]

\end{tabular}
}

\caption{\textbf{Examples of proposed dimensionality reduction-based saliency types across all explored biometric attack domains.} Each row provides a single sample and four distinct saliency maps from the evaluated datasets.}
\label{fig:saliency-types}
\end{figure*}

In order to explore the efficacy of dimensionality reduction-based saliency types inspired by Eigenfaces and Fisherfaces, we first produced the saliency maps in question. Once acquired, these saliency types guided the training of varying CNN architectures across five biometric attack detection domains: iris PAD, synthetic face detection, fingerprint PAD, fingerprint vein PAD, and identification document PAD.

\subsection{Datasets}

\subsubsection{Saliency-explored Domains}

For \textbf{iris PAD}, we follow splits used by prior saliency-guided training research in the task \cite{boyd2022human, crum2024grains}. The training and validation data originate from various live and presentation attack iris datasets \cite{casia-database,Sung_OE_2007,Galbally_ICB_2012,Kohli_ICB_2013,Yambay_ISBA_2017,Trokielewicz_IVC_2020,Kohli_BTAS_2016,Wei_ICPR_2008,Trokielewicz_BTAS_2015,Yambay_IJCB_2017,Das_IJCB_2020}. We test on the 2020 Iris Liveness Detection Competition testing set \cite{boyd2020iris}. All iris datasets are available for licensed use by their respective authors.

For \textbf{synthetic face detection}, we also follow splits used by prior saliency-guided training research in the task \cite{boyd2023cyborg, crum2024grains}. The training and validation data consist of samples from the FRGCSubset \cite{FRGC-subset}, SREFI \cite{srefi} and \textit{thispersondoesnotexist.com}-sourced StyleGan2 images. We test on a combination of various real \cite{karras2017progressive, original-Celeb} and synthetic \cite{karras2017progressive, stargan, styleGAN2, Karras2020ada, karras2021sg3} face sets. All face and synthetic face datasets are available for licensed use by their respective authors.

For \textbf{fingerprint PAD}, we continually follow splits used by prior saliency-guided training research in the domain \cite{webster2025saliency}. The training and validation data is constructed through the training and testing splits of Fingerprint Liveness Detection Competitions from 2015 through 2021 \cite{livdet15, livdet17, livdet19, livdet21}. We test on the Fingerprint Liveness Detection Competition 2021 testing set \cite{livdet21}. All LivDet Fingerprint datasets are available for licensed use by their respective authors.

\subsubsection{Saliency-novel Domains}

For \textbf{fingerprint vein PAD}, we apply the CandyFV \cite{candyfv} dataset. Due to relative ease of the domain's PAD task, we composed a difficult limited train-test split using only frames captured using the color camera and NIR-950nm illumination, selected for having the lowest average MSE between bonafide and presentation attack samples and therefore being a worst-case capture environment. We train and validate on bonafide and Level A attack samples, which are easier to produce. We test on bonafide and the more difficult Level B attack samples.

For \textbf{identification document PAD}, we follow the SIDTD \cite{webster2026psychophysical} framework building upon the MIDV-2020 \cite{id_document_pad} dataset, which performs realistic attacks on the synthetic samples from MIDV. This supports an environment where the MIDV-2020 and SIDTD samples are considered bonafide and presentation attacks, respectively. Over the 2222 ID scan samples available over both sets, we apply produce splits approximating 40\% train, 10\% validate, 50\% test with no identity leakage between splits.

\subsection{Generation of Saliency Types}
\label{subsec:saliency-types}

Using each domain's training data split alone, we explore four generated saliency types, termed and constructed as follows:

\begin{itemize}
    \item \textbf{Eigenface Saliency.} First, following the process introduced by Turk et al. \cite{eigenfaces}, we perform Principal Component Analysis on all available training samples, centered by the average training image ($img - img_{avg}$). The discovered components are the \textit{Eigenfaces}. By transforming a centered input image through the learned PCA, we produce a vector of Eigenface weights. This vector is then reverse transformed through the PCA, forming a weighted combination of all Eigenfaces that approximates the input image. To produce the final Eigenface saliency map, the absolute value is taken and then it is normalized to \{0-255\}. 
    
    \item \textbf{Eigenface Minus \% Fidelity (M-\%) Saliency.} A nearly identical process is taken as Eigenface saliency, except we progressively remove the top, most influential Eigenfaces identified by PCA until the MSE between the full-component Eigenface saliency exceeds a specified threshold. We compose "Eigenface M-\%" saliency at error percentages \{0.1, 0.2, 0.3, 0.4, 0.5, 0.6, 0.7, 0.8, 0.9\}. This saliency type is motivated by evidence where greater Eigenface facial recognition accuracy is achieved when omitting early, important components \cite{fisherfaces}. Since these components largely capture lighting, structural, and other non-distinguishing variances in images, their subtraction may benefit saliency-guided training, which we test at the nine articulated levels of reduced fidelity.
    
    \item \textbf{Fisherface Saliency.} Following the process introduced by Belhumeur \cite{fisherfaces}, we perform PCA on the training set as above methods. After this, Linear Discriminant Analysis is performed on the PCA-transformed data. The discovered scalings are the \textit{Fisherfaces}. To produce Fisherface saliency, an input image is projected first into PCA and then into LDA. The resulting vector is reverse transformed through LDA and then through PCA, producing a map denoting regions of class separability. It is identically processed with absolute value and normalization to \{0-255\}. 
    
    \item \textbf{LDA Saliency.} As Eigenface saliency represents PCA-only processing and Fisherface represents PCA and LDA processing combinatorially, we wanted to lastly explore saliency using LDA alone. To produce this saliency type, LDA is performed directly on the centered input images, unlike on the PCA-transformed input in Fisherface. Then, input images are transformed by LDA and reverse transformed into a weighted combination of LDA scalings, which are composed to create the final LDA saliency. The saliency is identically processed with absolute value and normalization to \{0-255\}. 
\end{itemize}

All saliency types are visually represented for all four domains explored in Fig. \ref{fig:saliency-types}.

\subsection{Saliency-guided Training Experiments}

Using the four described saliency types, we train saliency-guided models to explore their performance in key metrics. We evaluate and select top models based on AUC followed by classification accuracy at Equal Error Rate threshold (and 0.5 for fingerprint vein PAD due to task ease), metrics common in prior saliency-guided training works \cite{crum2024grains, webster2025saliency}. We additionally consider attack presentation classification error rate (APCER) and bonafide presentation classification error rate (BPCER), compliant with ISO/IEC 30107-3 on biometric attack detection testing \cite{ISO30107-3}. We specifically train CNN architectures ResNet50 \cite{resnet}, DenseNet-121 \cite{densenet}, and Inception-V3 \cite{inception}, based on prior saliency-guided experimentation on the same architectures. All configurations are trained three times for 50 epochs, using a batch size of 20 and a learning rate of 0.005 for a Stochastic Gradient Descent optimizer. For each configuration, we evaluate the model that achieved the lowest validation loss and average results over the three runs, reporting standard deviation when available. 

The CYBORG loss function \cite{boyd2023cyborg} has a sensitive $\alpha$ parameter weighing cross entropy and CAM-saliency map alignment functions. Past works in saliency-guided biometric attack detection have considered single fixed value ($\alpha = 0.5$) \cite{boyd2023cyborg, crum2024grains} or a set of fixed values ($\alpha=\{0.1, 0.3, 0.5, 0.7, 0.9\}$ \cite{webster2025saliency, webster2026psychophysical}. We opt to apply a successful tuning strategy which treats $\alpha$ as a learnable parameter, dynamically balancing it during training \cite{webster2026psychophysical}.

% Results
\section{Results}

\definecolor{baseline}{gray}{0.90}
\definecolor{sota}{RGB}{230,235,255}
\definecolor{best}{RGB}{230,255,230}

\begin{table*}
\centering
\caption{\textbf{A summary of proposed saliency performance across explored domains.} Rows \colorbox{sota}{highlighted in light blue} describe reported the state-of-the-art saliency in a domain if it exists. The highest-AUC-earning configuration is described for each saliency type in each domain; for Eigenface M-\% saliency, we list only the best percentile map in each domain. For each domain and excluding SOTA, the best AUC, classification accuracy, attack presentation classification error rate (APCER), and bonafide presentation classification error rate (BPCER) scores are \textbf{bolded}. The model with the best AUC (or accuracy, if a tie) is \colorbox{best}{highlighted in light green.} The best score exceeding SOTA, if any, is \underline{underlined}. Full training results are available in the Appendix.}
\label{tab:results-summary}

% \newcolumntype{L}{>{\raggedright\arraybackslash}X}
\newcolumntype{C}{>{\centering\arraybackslash}X}

\begin{tabularx}{\linewidth}{l | l l | C C C C}
\toprule
\textbf{Domain} & \textbf{Saliency Type} & \textbf{Architecture} & \textbf{AUC $\uparrow$} & \textbf{Accuracy $\uparrow$} & \textbf{APCER $\downarrow$} & \textbf{BPCER $\downarrow$} \\
\midrule

% --- Iris PAD ---
\multirow{5}{*}{\textbf{Iris PAD}} 
& \cellcolor{baseline}None & \cellcolor{baseline}DenseNet & \cellcolor{baseline}0.896$\pm$0.014 & \cellcolor{baseline}0.814$\pm$0.011 & \cellcolor{baseline}0.151$\pm$0.024 & \cellcolor{baseline}\textbf{0.234$\pm$0.043} \\
& \cellcolor{sota}Autoencoder \cite{crum2024grains} & \cellcolor{sota}DenseNet & \cellcolor{sota}\textbf{0.962$\pm$0.005} & \cellcolor{sota}- & \cellcolor{sota}- & \cellcolor{sota}- \\
& Eigenface & Inception & 0.872$\pm$0.023 & 0.784$\pm$0.023 & 0.117$\pm$0.017 & 0.348$\pm$0.038 \\
& Eigenface M-0.2 & ResNet & 0.887$\pm$0.008 & 0.803$\pm$0.008 & 0.140$\pm$0.004 & 0.273$\pm$0.022 \\
& Fisherface & ResNet & 0.896$\pm$0.003 & 0.809$\pm$0.002 & 0.108$\pm$0.028 & 0.301$\pm$0.032 \\
& \cellcolor{best}\textbf{LDA} & \cellcolor{best}\textbf{DenseNet} & \cellcolor{best}0.911$\pm$0.012 & \cellcolor{best}\textbf{0.824$\pm$0.012} & \cellcolor{best}\textbf{0.102$\pm$0.020} & \cellcolor{best}0.275$\pm$0.026 \\

\midrule

% --- Synthetic Face Detection ---
\multirow{5}{*}{\parbox{2cm}{\textbf{Synthetic Face\\Detection}}} 
& \cellcolor{baseline}None & \cellcolor{baseline}Inception & \cellcolor{baseline}0.620$\pm$0.022 & \cellcolor{baseline}0.848$\pm$0.008 & \cellcolor{baseline}0.014$\pm$0.010 & \cellcolor{baseline}0.983$\pm$0.006\\ 
& \cellcolor{sota}Human Annotation \cite{crum2024grains} & \cellcolor{sota}DenseNet & \cellcolor{sota}0.643$\pm$0.033 & \cellcolor{sota}- &  \cellcolor{sota}- &  \cellcolor{sota}-\\
& Eigenface & Inception & 0.598$\pm$0.050 & \textbf{0.856$\pm$0.001} & \textbf{0.005$\pm$0.002} & 0.978$\pm$0.008\\
& Eigenface M-0.4 & Inception & 0.627$\pm$0.020 & 0.854$\pm$0.003 & 0.009$\pm$0.004 & 0.972$\pm$0.010 \\
& \cellcolor{best}\textbf{Fisherface} & \cellcolor{best}\textbf{Inception} & \cellcolor{best}\textbf{\underline{0.654$\pm$0.051}} & \cellcolor{best}0.854$\pm$0.003 & \cellcolor{best}0.008$\pm$0.004 & \cellcolor{best}\textbf{0.970$\pm$0.023} \\
& LDA & Inception & \textbf{\underline{0.654$\pm$0.003}} & 0.853$\pm$0.003 & 0.010$\pm$0.008 & 0.971$\pm$0.023 \\

\midrule

% --- Fingerprint PAD ---
\multirow{5}{*}{\parbox{2cm}{\textbf{Fingerprint\\PAD}}} 
& \cellcolor{baseline}None & \cellcolor{baseline}Inception & \cellcolor{baseline}0.946$\pm$0.007 & \cellcolor{baseline}0.862$\pm$0.010 & \cellcolor{baseline}0.231$\pm$0.014 & \cellcolor{baseline}0.052$\pm$0.009 \\
& \cellcolor{sota}Minutiae Software \cite{webster2025saliency} & \cellcolor{sota}Inception & \cellcolor{sota}0.961$\pm$0.004 & \cellcolor{sota}0.885$\pm$0.004 & \cellcolor{sota}0.188$\pm$0.014 & \cellcolor{sota}0.048$\pm$0.006 \\
& Eigenface & Inception & 0.957$\pm$0.006 & 0.875$\pm$0.009 & 0.215$\pm$0.016 & \textbf{\underline{0.043$\pm$0.006}} \\
& \cellcolor{best}\textbf{Eigenface M-0.1} & \cellcolor{best}\textbf{Inception} & \cellcolor{best}\textbf{\underline{0.962$\pm$0.001}} & \cellcolor{best}\textbf{0.885$\pm$0.003} & \cellcolor{best}\textbf{0.188$\pm$0.009} & \cellcolor{best}0.048$\pm$0.004 \\
& Fisherface & Inception & 0.957$\pm$0.005 & 0.877$\pm$0.001 & 0.206$\pm$0.011 & 0.046$\pm$0.012 \\
& LDA & Inception & \textbf{\underline{0.962$\pm$0.002}} & 0.884$\pm$0.003 & 0.194$\pm$0.009 & 0.045$\pm$0.003\\

\toprule

\multirow{4}{*}{\parbox{2cm}{\textbf{Finger Vein \\ PAD}}}
& \cellcolor{baseline}None & \cellcolor{baseline}Densenet & \cellcolor{baseline}1.000$\pm$0.000 & \cellcolor{baseline}0.927$\pm$0.044 & \cellcolor{baseline}0.099$\pm$0.060 & \cellcolor{baseline}0.000$\pm$0.000 \\
& Eigenface & ResNet & 1.000$\pm$0.000 & 0.921$\pm$0.010 & 0.107$\pm$0.013 & 0.000$\pm$0.000 \\
& \cellcolor{best}\textbf{Eigenface M-0.1} & \cellcolor{best}\textbf{Inception} & \cellcolor{best}1.000$\pm$0.000 & \cellcolor{best}\textbf{0.999$\pm$0.001} & \cellcolor{best}\textbf{0.000$\pm$0.000} & \cellcolor{best}0.006$\pm$0.004 \\
& Fisherface & Inception & 1.000$\pm$0.000 & 0.930$\pm$0.033 & 0.096$\pm$0.045 & 0.000$\pm$0.000 \\
& LDA & Inception & 1.000$\pm$0.000 & 0.998$\pm$0.002 & 0.002$\pm$0.002 & 0.000$\pm$0.000\\

\midrule

\multirow{4}{*}{\parbox{2cm}{\textbf{Identification \\ Document \\ PAD}}}
& \cellcolor{baseline}None & \cellcolor{baseline}Inception & \cellcolor{baseline}0.884$\pm$0.072 & \cellcolor{baseline}0.812$\pm$0.055 & \cellcolor{baseline}0.210$\pm$0.065 & \cellcolor{baseline}0.157$\pm$0.054 \\
& Eigenface & DenseNet & 0.929$\pm$0.015 & 0.842$\pm$0.013 & 0.172$\pm$0.011 & 0.138$\pm$0.028 \\
& Eigenface M-0.4 & DenseNet & 0.923$\pm$0.016 & 0.845$\pm$0.020 & 0.166$\pm$0.026 & 0.140$\pm$0.036 \\
& Fisherface & DenseNet &  0.940$\pm$0.030 & 0.862$\pm$0.035 & 0.173$\pm$0.032 & \textbf{0.091$\pm$0.039} \\
& \cellcolor{best}\textbf{LDA} & \cellcolor{best}\textbf{DenseNet} & \cellcolor{best}\textbf{0.950$\pm$0.014} & \cellcolor{best}\textbf{0.873$\pm$0.023} & \cellcolor{best}\textbf{0.148$\pm$0.035} & \cellcolor{best}0.099$\pm$0.011 \\

\bottomrule
\end{tabularx}
\end{table*}

The results of our saliency-guided training experimentation are summarized in Tab. \ref{tab:results-summary}. We further examine these results by directly answering our research questions.

\subsection{RQ1: Are dimensionality reduction-sourced saliency maps competitive for guiding biometric PAD models compared to existing methods?}

In \textbf{iris PAD}, the proposed LDA-based saliency maps exceed the best cross entropy baseline AUC, accuracy, and APCER scores, making them a viable option for saliency-guided training. However, this configuration's AUC performance is 0.051 less than the current state-of-the-art (SOTA) saliency type: sourced from an autoencoder model trained to mimic human annotators \cite{crum2024grains}. In this domain, Fisherface-based saliency maps roughly match AUC and accuracy performance, sacrificing bonafide accuracy for improvements to APCER. Both Eigenface-based saliency approaches fail to exceed the baseline metrics. To address RQ1, \textbf{yes}: in iris PAD saliency maps based in classical dimensionality reduction are capable of guiding models toward improved performance, but their gains \textbf{do not exceed SOTA methods}.

In \textbf{synthetic face detection}, the best-performing Eigenface M-0.4, Fisherface, and LDA-based saliency types all exceed the best baseline model over all metrics by similar improvements. Despite not exceeding the AUC baseline, Eigenface-based saliency narrowly achieves the highest accuracy within the domain. Further, both Fisherface and LDA saliency types reach an AUC of 0.654, which exceeds the current human-annotative saliency SOTA \cite{crum2024grains} by 0.011. Answering the research question, \textbf{yes}: in synthetic face detection, the proposed saliency strategies are capable of exceeding not only baseline models but also the current SOTA guidance method.

In \textbf{fingerprint PAD}, all proposed saliency types exceed the baseline performance in every metric, with Eigenface-M1 and LDA saliency acquisition methods realizing near-identical gains to AUC and accuracy. These methods eke past the SOTA software-sourced minutiae maps \cite{webster2025saliency} in AUC; additionally, the Eigenface saliency approach reduces BPCER lower than the SOTA method. While the improvements are statistically equal, we can answer \textbf{yes}, in fingeprint PAD, dimensionality reduction-sourced saliency maps are able to exceed baseline and marginally SOTA performance metrics.

\subsection{RQ2: Are dimensionality reduction-sourced saliency maps effective in saliency-novel biometric PAD domains?}

The \textbf{fingerprint vein PAD} was trivial for our models, with baseline and all variants of our proposed saliency methods achieving an AUC of 1.0 as well as near 0.0 error in bonafide classification. Consequently, an optimal threshold produces perfect accuracy and no error. Thus, we assess the fingerprint vein PAD task at a threshold of 0.5. At this operating point, the Eigenface M-0.1-, Fisherface-, LDA-guided models exceed baseline accuracy and APCER scores. Both Eigenface M-0.1 and LDA-based saliency types achieve a near perfect accuracy and exceed the baseline by ~7\% at this threshold, showcasing the power of saliency-guided training as a regularizer. To answer RQ2 from the fingerprint vein PAD lens, \textbf{yes}, the proposed dimensionality reduction-sourced saliency maps can effectively generalize to domains unexplored by saliency-guided training.

In \textbf{identification document PAD}, Eigenface, Eigenface M-0.4, Fisherface, and LDA saliency strategies all produce notable improvements over optimal baseline performance in every metric. The LDA-based saliency type demonstrates the greatest performance, improving AUC by 6.6\% and accuracy by 6.1\% while reducing APCER by 6.2\%. Fisherface-based saliency achieves the lowest BPCER at 0.091. To again address RQ2, \textbf{yes}, the proposed saliency acquisition strategies are effective in identification document PAD despite its lack of exploration in saliency guidance.

\subsection{Ablation Study: Varying Eigenface M-\%}
\label{subsec:ablation}

\begin{figure}
\centering

\includegraphics[width=0.45\textwidth]{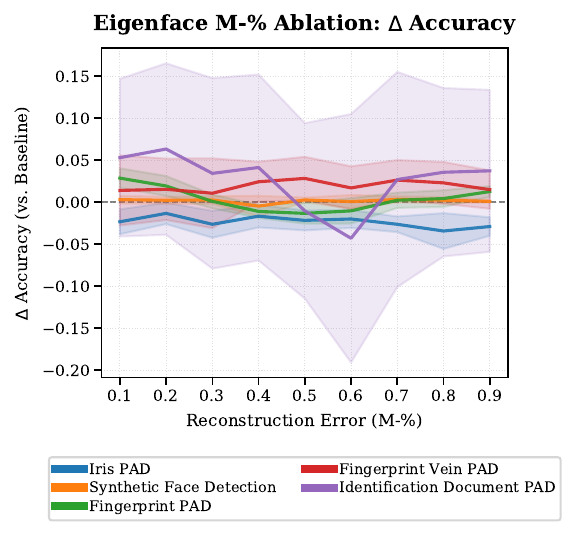}
\caption{\textbf{Changes to accuracy when varying Eigenface M-\% across all biometric attack detection domains.} Each domain's solid line denotes mean and transparent region denotes $\pm1$ standard deviation. There is a subtle global preference toward lower error percentiles and a negative reaction to mid-range values in ID Document and Fingerprint Vein PAD.}
\label{fig:ablation}
\end{figure}

As described in Sec. \ref{subsec:saliency-types}, we perform the Eigenface M-\% saliency process at 9 levels of reconstruction error between 0.1 and 0.9. Rather than omitting an arbitrary sum of principal components, this ablation study allows us to systematically apply the observations related to Eigenface top component omission to our saliency acquisition process \cite{fisherfaces}. Tab. \ref{tab:omitted} summarizes the average number of principal components omitted to reach each error percentile, aggregated within each domain and across all domains. The foundational nature of top components is showcased; across all domains, the average components omitted to reach 20\% error or less is under 6--not considering fingerprint PAD, it takes on average less than 9 omitted components to reach up to 40\% error. Conversely, for higher reconstruction error percentiles, the average number of components required as well as the gap of components between percentiles increase dramatically, demonstrating the diminishing influence of components in PCA.

\begin{table*}
\centering
\caption{The average number of top PCA components removed to reach each Eigenface M-\% error percentile over all domains.}
\label{tab:omitted}

\newcolumntype{L}{>{\raggedright\arraybackslash}X}
\newcolumntype{C}{>{\centering\arraybackslash}X}

\small
\begin{tabularx}{\linewidth}{l | C C C C C C C C C }
\toprule

\rowcolor{header}
Domain & M-0.1 & M-0.2 & M-0.3 & M-0.4 & M-0.5 & M-0.6 & M-0.7 & M-0.8 & M-0.9 \\
\midrule

Iris PAD & 1.91 & 3.04 & 4.90 & 9.15 & 19.47 & 45.01 & 104.87 & 241.14 & 462.78 \\
Synthetic Face Detection & 2.29 & 4.18 & 7.22 & 13.96 & 30.57 & 70.87 & 176.94 & 461.21 & 1059.45 \\
Fingerprint PAD & 3.01 & 16.30 & 80.03 & 202.02 & 322.71 & 397.74 & 447.95 & 492.90 & 567.98 \\
Fingerprint Vein PAD & 1.28 & 1.52 & 2.10 & 3.69 & 7.89 & 17.06 & 33.01 & 74.55 & 266.86 \\
Identification Card PAD & 2.16 & 3.14 & 4.75 & 7.08 & 12.38 & 35.05 & 140.64 & 317.66 & 531.67 \\
\midrule
All Domains & 2.13 & 5.63 & 19.80 & 47.18 & 78.60 & 113.15 & 180.68 & 317.49 & 577.75 \\

\bottomrule
\end{tabularx}
\end{table*}

Figure \ref{fig:ablation} summarizes metric changes compared to baseline for Eigenface M-\% percentiles; full ablation results are available in the Appendix. The effect of varying M-\% is dependent on domain. In fingerprint, fingerprint vein, and identification card PAD tasks, a slight preference for the lower M-\% maps is observed; this is validated by Eigenface M-0.1 saliency achieving the highest in-domain performance for fingerprint and fingerprint vein PAD. Identification document PAD is most sensitive to change in M-\%, exhibiting a strong negative change to performance around the 0.6 error percentile. 

While the Eigenface M-\% saliency maps are not consistently more effective than the proposed Fisherface and LDA-based strategies, they outperform full-component Eigenface maps in every domain tested. This ablation study affirms when guiding biometric attack detection models with Eigenface-inspired saliency maps, removing top components from Eigenface signatures dependably exceeds their original version, closely paralleling the benefits realized in face recognition observed by the same process \cite{fisherfaces}.

% Discussion
\section{Discussion}

This work introduces a novel saliency acquisition method inspired by classical dimensionality reduction techniques. We demonstrate our method's prerequisite-free adoption and competitive gains to accuracy and generalization in five biometric attack detection tasks, exceeding the state-of-the-art saliency implementations in synthetic face detection and fingerprint PAD. By bypassing steep resource barriers exhibited by current saliency map acquisition methods, our strategy offers a path toward efficient saliency-guided training, supporting robust training in new domains. 

\subsection{Effectiveness of Proposed Saliency Methods}

Our experimental results indicate that dimensionality reduction-sourced saliency benefits from LDA processing--namely, the LDA and Fisherface saliency sources. Ranking by achieved AUC and accuracy, LDA-based saliency maps perform the best in three tasks and second-best in two tasks among the proposed saliency types. Continually, Fisherface saliency performs the best in one domain and second best in two. This suggests that LDA-influenced maps, which annotate class-discriminating directions, are generally more effective as guiding saliency in for training biometric presentation attack detection models. 

Despite being occasionally outperformed by Fisherface and LDA saliency types, both Eigenface variants were effective. The Eigenface M-\% saliency maps achieved the highest performance in two explored domains. Pure Eigenface saliency is the lowest-performing saliency style proposed; in four tasks, it is matched or outperformed by another proposed saliency type. This is behavior is perhaps attributable to the lack of 'newness' presented by full-component Eigenface maps, which effectively compose the input image minus the dataset average. The contrasting success of Eigenface M-\% saliency reinforces this notion, as the process of removing structural and common features enables the M-\% saliency maps to communicate unique and statistically variant image regions. 

\subsection{Efficient Saliency-guided Training}

A key contribution of our dimensionality reduction-based saliency acquisition methods are their efficient and low-prerequisite implementation, supporting straightforward saliency-guided implementation for new biometric domains. While existing saliency methods have been critically limited on expensive collection processes as well as existence of applied technology or annotations, as summarized in Tab. \ref{tab:saliency-methods}, the proposed maps rely only on the training data being annotated. This characteristic democratizes robust saliency-guided training for biometric domains where limited resources or a lack of domain tooling may not support existing saliency methods.

\subsection{Limitations and Future Work}

As dimensionality reduction techniques identify directions of variance within a dataset or class, they are limited to the statistical variations present in the provided dataset. If a dataset whose variance is insufficient for describing class-relative features through PCA or LDA, then the generated maps may yield poor results in saliency-guided training. Conversely, a biometric dataset with class-identifying bias, such as capture device or lighting, may aggressively guide toward these weakly-generalizing dataset-specific signals.

Future work may seek to mitigate this limitation, such as a component filtering scheme based on unexpectedly high class discrimination, suggesting bias or label leakage. Modifying the proposed saliency framework to use Robust Principal Component Analysis \cite{netrapalli2014non} or Robust Discriminant Analysis \cite{hubert2024robust} may further support generalizable saliency maps for unrepresentative or skewed datasets.

{\small
\bibliographystyle{ieee}
\bibliography{egbib}

@inproceedings{crum2024grains,
  title={Grains of Saliency: Optimizing Saliency-based Training of Biometric Attack Detection Models},
  author={Crum, Colton R and Webster, Samuel and Czajka, Adam},
  booktitle={2024 IEEE International Joint Conference on Biometrics (IJCB)},
  pages={1--9},
  year={2024},
  organization={IEEE}
}

@inproceedings{boyd2022human,
  title={Human-aided saliency maps improve generalization of deep learning},
  author={Boyd, Aidan and Bowyer, Kevin W and Czajka, Adam},
  booktitle={Proceedings of the IEEE/CVF Winter Conference on Applications of Computer Vision},
  pages={2735--2744},
  year={2022}
}

@inproceedings{boyd2023cyborg,
  title={Cyborg: Blending human saliency into the loss improves deep learning-based synthetic face detection},
  author={Boyd, Aidan and Tinsley, Patrick and Bowyer, Kevin W and Czajka, Adam},
  booktitle={Proceedings of the IEEE/CVF Winter Conference on Applications of Computer Vision},
  pages={6108--6117},
  year={2023}
}

@INPROCEEDINGS{webster2025saliency,
  author={Webster, Samuel and Czajka, Adam},
  booktitle={2025 IEEE International Joint Conference on Biometrics (IJCB)}, 
  title={Saliency-Guided Training for Fingerprint Presentation Attack Detection}, 
  year={2025},
  volume={},
  number={},
  pages={1-10},
  doi={10.1109/IJCB65343.2025.11411553}
}

@ARTICLE{webster2026psychophysical,
  author={Webster, Samuel and Scheirer, Walter and Czajka, Adam},
  journal={IEEE Transactions on Biometrics, Behavior, and Identity Science}, 
  title={Psychophysically-Guided Training for Fingerprint Presentation Attack Detection}, 
  year={2026},
  volume={},
  number={},
  pages={1-1},
  doi={10.1109/TBIOM.2026.3664679}
}

@inproceedings{eigenfaces,
  title={Face recognition using eigenfaces.},
  author={Turk, Matthew A and Pentland, Alex and others},
  booktitle={CVPR},
  volume={91},
  pages={586--591},
  year={1991}
}

@article{fisherfaces,
  title={Eigenfaces vs. fisherfaces: Recognition using class specific linear projection},
  author={Belhumeur, Peter N. and Hespanha, Joao P and Kriegman, David J.},
  journal={IEEE Transactions on pattern analysis and machine intelligence},
  volume={19},
  number={7},
  pages={711--720},
  year={1997},
  publisher={IEEE}
}

@misc{resnet,
  title={Deep residual learning for image recognition. CoRR abs/1512.03385 (2015)},
  author={He, Kaiming and Zhang, Xiangyu and Ren, Shaoqing and Sun, Jian},
  year={2015}
}

@inproceedings{densenet,
  title={Densely connected convolutional networks},
  author={Huang, Gao and Liu, Zhuang and Van Der Maaten, Laurens and Weinberger, Kilian Q},
  booktitle={Proceedings of the IEEE conference on computer vision and pattern recognition},
  pages={4700--4708},
  year={2017}
}

@inproceedings{inception,
  title={Inception-v4, inception-resnet and the impact of residual connections on learning},
  author={Szegedy, Christian and Ioffe, Sergey and Vanhoucke, Vincent and Alemi, Alexander},
  booktitle={Proceedings of the AAAI conference on artificial intelligence},
  volume={31},
  year={2017}
}

@article{boyd2020iris,
  title={Iris presentation attack detection: Where are we now?},
  author={Boyd, Aidan and Fang, Zhaoyuan and Czajka, Adam and Bowyer, Kevin W},
  journal={Pattern Recognition Letters},
  volume={138},
  pages={483--489},
  year={2020},
  publisher={Elsevier}
}

@misc{casia-database,
  title = {{Chinese Academy of Sciences Institute of Automation}},
  howpublished = {http://www.cbsr.ia.ac.cn/china/Iris\%20Databases\%20CH.asp},
  note = {Accessed: 03-12-2021}
}

@inproceedings{Kohli_ICB_2013,
    title={Revisiting iris recognition with color cosmetic contact lenses},
  author={Kohli, Naman and Yadav, Daksha and Vatsa, Mayank and Singh, Richa},
  booktitle={2013 International Conference on Biometrics (ICB)},
  pages={1--7},
  year={2013},
  organization={IEEE}
}

@inproceedings{Galbally_ICB_2012,
    author    = {Javier Galbally and Jaime Ortiz-Lopez and Julian Fierrez and Javier Ortega-Garcia},
    title     = {Iris liveness detection based on quality related features},
    booktitle = {2012 5th IAPR Int. Conf. on Biometrics (ICB)},
    address   = {New Delhi, India},
    publisher = {IEEE},
    year      = {2012},
    pages     = {271-276},
    month     = {March},
    comment   = {iris:PAD:static:passive},
    doi       = {10.1109/ICB.2012.6199819},
    issn      = {2376-4201},
}

@inproceedings{Wei_ICPR_2008,
    title={Synthesis of large realistic iris databases using patch-based sampling},
  author={Wei, Zhuoshi and Tan, Tieniu and Sun, Zhenan},
  booktitle={2008 19th International Conference on Pattern Recognition},
  pages={1--4},
  year={2008},
  organization={IEEE}
}

@inproceedings{Kohli_BTAS_2016,
    title={Detecting medley of iris spoofing attacks using DESIST},
  author={Kohli, Naman and Yadav, Daksha and Vatsa, Mayank and Singh, Richa and Noore, Afzel},
  booktitle={2016 IEEE 8th International Conference on Biometrics Theory, Applications and Systems (BTAS)},
  pages={1--6},
  year={2016},
  organization={IEEE}
}

@article{Trokielewicz_IVC_2020,
    title    = {Post-mortem iris recognition with deep-learning-based image segmentation},
    journal  = {Image and Vision Computing},
    volume   = {94},
    pages    = {103866},
    year     = {2020},
    issn     = {0262-8856},
    doi      = {https://doi.org/10.1016/j.imavis.2019.103866},
    url      = {http://www.sciencedirect.com/science/article/pii/S0262885619304597},
    author   = {Mateusz Trokielewicz and Adam Czajka and Piotr Maciejewicz}
}

@article{Sung_OE_2007,
    author    = {Sung Joo Lee and Kang Ryoung Park and Youn Joo Lee and Kwanghyuk Bae and Jai Hie Kim},
    title     = {{Multifeature-based fake iris detection method}},
    volume    = {46},
    journal   = {Optical Engineering},
    number    = {12},
    publisher = {SPIE},
    pages     = {1 -- 10},
    year      = {2007},
    doi       = {10.1117/1.2815719},
    url       = {https://doi.org/10.1117/1.2815719},
}

@inproceedings{Yambay_ISBA_2017,
   title={LivDet iris 2017—Iris liveness detection competition 2017},
  author={Yambay, David and Becker, Benedict and Kohli, Naman and Yadav, Daksha and Czajka, Adam and Bowyer, Kevin W and Schuckers, Stephanie and Singh, Richa and Vatsa, Mayank and Noore, Afzel and others},
  booktitle={2017 IEEE International Joint Conference on Biometrics (IJCB)},
  pages={733--741},
  year={2017},
  organization={IEEE}
}

@InProceedings{Yambay_IJCB_2017,
  title={LivDet iris 2017—Iris liveness detection competition 2017},
  author={Yambay, David and Becker, Benedict and Kohli, Naman and Yadav, Daksha and Czajka, Adam and Bowyer, Kevin W and Schuckers, Stephanie and Singh, Richa and Vatsa, Mayank and Noore, Afzel and others},
  booktitle={2017 IEEE International Joint Conference on Biometrics (IJCB)},
  pages={733--741},
  year={2017},
  organization={IEEE}
}

@inproceedings{Trokielewicz_BTAS_2015,
    title={Assessment of iris recognition reliability for eyes affected by ocular pathologies},
  author={Trokielewicz, Mateusz and Czajka, Adam and Maciejewicz, Piotr},
  booktitle={2015 IEEE 7th International Conference on Biometrics Theory, Applications and Systems (BTAS)},
  pages={1--6},
  year={2015},
  organization={IEEE}
}

@INPROCEEDINGS{Das_IJCB_2020,
  author={P. {Das} and J. {Mcfiratht} and Z. {Fang} and A. {Boyd} and G. {Jang} and A. {Mohammadi} and S. {Purnapatra} and D. {Yambay} and S. {Marcel} and M. {Trokielewicz} and P. {Maciejewicz} and K. {Bowyer} and A. {Czajka} and S. {Schuckers} and J. {Tapia} and S. {Gonzalez} and M. {Fang} and N. {Damer} and F. {Boutros} and A. {Kuijper} and R. {Sharma} and C. {Chen} and A. {Ross}},
  booktitle={2020 IEEE International Joint Conference on Biometrics (IJCB)}, 
  title={{Iris Liveness Detection Competition (LivDet-Iris) - The 2020 Edition}}, 
  year={2020},
  volume={},
  number={},
  pages={1-9},
  doi={10.1109/IJCB48548.2020.9304941}
}

@article{FRGC-subset,
  title={Lessons from collecting a million biometric samples},
  author={Phillips, P Jonathon and Flynn, Patrick J and Bowyer, Kevin W},
  journal={Image and Vision Computing},
  volume={58},
  pages={96--107},
  year={2017},
  publisher={Elsevier}
}

@inproceedings{original-Celeb,
  title={Deep learning face attributes in the wild},
  author={Liu, Ziwei and Luo, Ping and Wang, Xiaogang and Tang, Xiaoou},
  booktitle={Proceedings of the IEEE international conference on computer vision},
  pages={3730--3738},
  year={2015}
}

@inproceedings{srefi,
  title={Srefi: Synthesis of realistic example face images},
  author={Banerjee, Sandipan and Bernhard, John S and Scheirer, Walter J and Bowyer, Kevin W and Flynn, Patrick J},
  booktitle={2017 IEEE International Joint Conference on Biometrics (IJCB)},
  pages={37--45},
  year={2017},
  organization={IEEE}
}

@inproceedings{stargan,
  title={Stargan v2: Diverse image synthesis for multiple domains},
  author={Choi, Yunjey and Uh, Youngjung and Yoo, Jaejun and Ha, Jung-Woo},
  booktitle={Proceedings of the IEEE/CVF conference on computer vision and pattern recognition},
  pages={8188--8197},
  year={2020}
}

@inproceedings{styleGAN2,
  title={Analyzing and improving the image quality of stylegan},
  author={Karras, Tero and Laine, Samuli and Aittala, Miika and Hellsten, Janne and Lehtinen, Jaakko and Aila, Timo},
  booktitle={Proceedings of the IEEE/CVF conference on computer vision and pattern recognition},
  pages={8110--8119},
  year={2020}
}

@article{karras2017progressive,
  title={{Progressive Growing of GANs for Improved Quality, Stability, and Variation}},
  author={Karras, Tero and Aila, Timo and Laine, Samuli and Lehtinen, Jaakko},
  journal={arXiv preprint arXiv:1710.10196},
  year={2017}
}

@inproceedings{Karras2020ada,
  title     = {Training Generative Adversarial Networks with Limited Data},
  author    = {Tero Karras and Miika Aittala and Janne Hellsten and Samuli Laine and Jaakko Lehtinen and Timo Aila},
  booktitle = {Proc. NeurIPS},
  year      = {2020}
}

@article{karras2021sg3,
  author = {Tero Karras and Miika Aittala and Samuli Laine and Erik H\"ark\"onen and Janne Hellsten and Jaakko Lehtinen and Timo Aila},
  title = {Alias-Free Generative Adversarial Networks},
  journal = {Proc. NeurIPS},
  year = {2021}
}

@inproceedings{livdet15,
  author={Mura, Valerio and Ghiani, Luca and Marcialis, Gian Luca and Roli, Fabio and Yambay, David A. and Schuckers, Stephanie A.},
  booktitle={2015 IEEE 7th International Conference on Biometrics Theory, Applications and Systems (BTAS)}, 
  title={LivDet 2015 fingerprint liveness detection competition 2015}, 
  year={2015},
  volume={},
  number={},
  pages={1-6},
  doi={10.1109/BTAS.2015.7358776}
}

@inproceedings{livdet17,
  title={LivDet 2017 fingerprint liveness detection competition 2017},
  author={Mura, Valerio and Orr{\`u}, Giulia and Casula, Roberto and Sibiriu, Alessandra and Loi, Giulia and Tuveri, Pierluigi and Ghiani, Luca and Marcialis, Gian Luca},
  booktitle={2018 international conference on biometrics (ICB)},
  pages={297--302},
  year={2018},
  organization={IEEE}
}

@inproceedings{livdet19,
  title={Livdet in action-fingerprint liveness detection competition 2019},
  author={Orr{\`u}, Giulia and Casula, Roberto and Tuveri, Pierluigi and Bazzoni, Carlotta and Dessalvi, Giovanna and Micheletto, Marco and Ghiani, Luca and Marcialis, Gian Luca},
  booktitle={2019 international conference on biometrics (ICB)},
  pages={1--6},
  year={2019},
  organization={IEEE}
}

@inproceedings{livdet21,
  title={Livdet 2021 fingerprint liveness detection competition-into the unknown},
  author={Casula, Roberto and Micheletto, Marco and Orr{\`u}, Giulia and Delussu, Rita and Concas, Sara and Panzino, Andrea and Marcialis, Gian Luca},
  booktitle={2021 IEEE international joint conference on biometrics (IJCB)},
  pages={1--6},
  year={2021},
  organization={IEEE}
}

@article{crum2023teaching,
  title={Teaching ai to teach: Leveraging limited human salience data into unlimited saliency-based training},
  author={Crum, Colton R and Boyd, Aidan and Bowyer, Kevin and Czajka, Adam},
  journal={arXiv preprint arXiv:2306.05527},
  year={2023}
}

@inproceedings{candyfv,
  author = {Bhattacharjee, Sushil and Geissbuhler, David and Clivaz, G. and Kotwal, Ketan and Marcel, S{\'{e}}bastien},
  title = {Vascular Biometrics Experiments on Candy -- A New Contactless Finger-Vein Dataset},
  booktitle = {Proceedings of the International Conference on Pattern Recognition (ICPR)},
  year = {2025},
  month = {Dec},
  }

@article{id_document_pad,
  title={MIDV-2020: A comprehensive benchmark dataset for identity document analysis},
  author={Bulatov, Konstantin and Emelianova, Ekaterina and Tropin, Daniil and Skoryukina, Natalya and Chernyshova, Yulia and Ming, Zuheng and Burie, Jean-Christophe and Luqman, Muhammad Muzzamil},
  journal={Computer Optics},
  volume={46},
  number={2},
  pages={252--270},
  year={2022},
  publisher={IPSI RAS}
}

@article{id_document_pad_2,
  title={Synthetic dataset of id and travel documents},
  author={Boned, Carlos and Talarmain, Maxime and Ghanmi, Nabil and Chiron, Guillaume and Biswas, Sanket and Awal, Ahmad Montaser and Ramos Terrades, Oriol},
  journal={Scientific data},
  volume={11},
  number={1},
  pages={1356},
  year={2024},
  publisher={Nature Publishing Group UK London}
}

@techreport{ISO30107-3,
    type = {Standard},
    key = {ISO/IEC 30107-3},
    year = {2023},
    title = {{Information technology -- Biometric presentation attack detection -- Testing and reporting}},
    volume = {2023},
    address = {Geneva, CH},
    institution = {International Organization for Standardization}
}

@article{netrapalli2014non,
  title={Non-convex robust PCA},
  author={Netrapalli, Praneeth and Niranjan, U N and Sanghavi, Sujay and Anandkumar, Animashree and Jain, Prateek},
  journal={Advances in neural information processing systems},
  volume={27},
  year={2014}
}

@article{hubert2024robust,
  title={Robust discriminant analysis},
  author={Hubert, Mia and Raymaekers, Jakob and Rousseeuw, Peter J},
  journal={Wiley Interdisciplinary Reviews: Computational Statistics},
  volume={16},
  number={5},
  pages={e70003},
  year={2024},
  publisher={Wiley Online Library}
}
}

% Temp: Supplementary Material
\newpage
\onecolumn

% \begin{center}
%     \LARGE \textbf{Supplementary Material for \\ ``What’s Old is New Again: Classical Dimensionality Reduction for Efficient Saliency-Guided Biometric Attack Detection''} \\[0.5em]
%     % \large Samuel Webster \& Walter Scheirer \\[0.3em]
%     \large Anonymous IJCB 2026 Submission
% \end{center}
% \vspace{1.5em}

\section*{Appendix}

\setcounter{section}{0}

\section{Supplementary Eigenface M-\% Ablation Charts}

\begin{figure}[H]
\centering

\begin{tabular}{c c c}

\toprule
\rowcolor{header}
\textbf{AUC} & \textbf{APCER} & \textbf{BPCER} \\
\bottomrule

\includegraphics[width=0.30\textwidth]{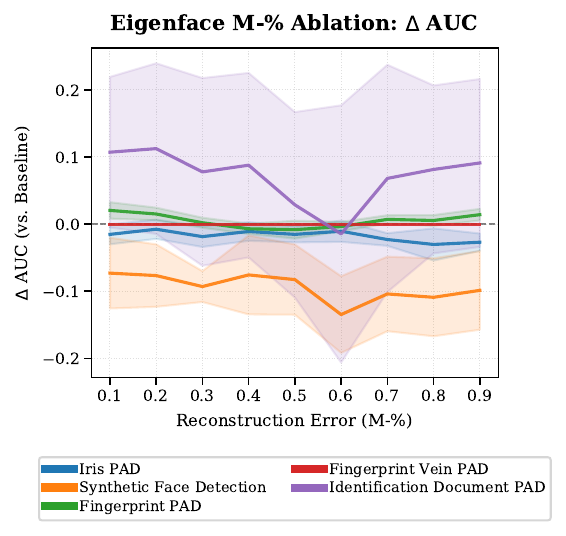} &
\includegraphics[width=0.30\textwidth]{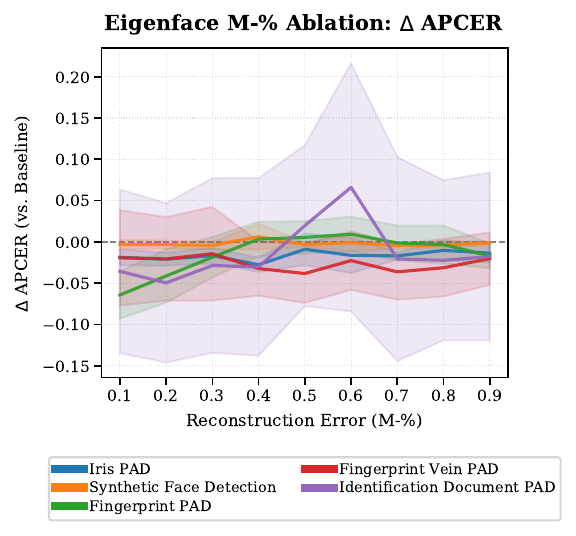} &
\includegraphics[width=0.30\textwidth]{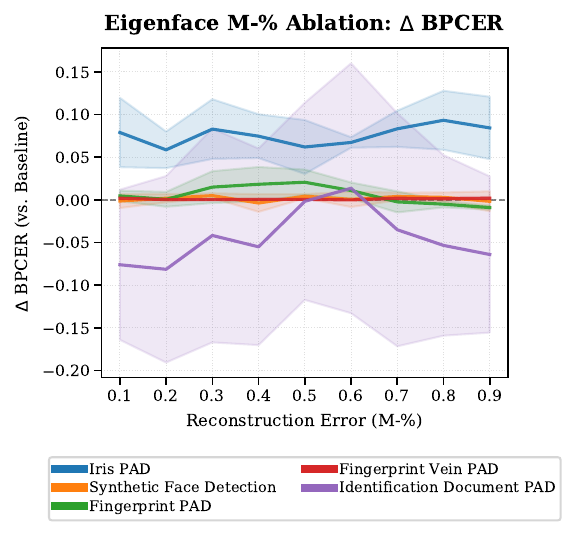} \\

\end{tabular}

\caption{\textbf{Changes to AUC, APCER, and BPCER when varying Eigenface M-\% across all biometric attack detection domains.}}
\end{figure}

Due to space constraints and limited expressivity of AUC and BPCER in Fingerprint Vein PAD, we prioritized inclusion of the Accuracy figure in the main paper. However, AUC, APCER, and BPCER share the global preference for lower Eigenface M-\% percentile errors as well as a substantial negative reaction in the mid-range maps, emphasized by Identification Document PAD and Synthetic Face Detection tasks.

\section{Unabridged Training Results}

\noindent All configurations denoted as `Baseline' and having gray rows are trained using cross entropy loss and without any saliency guidance, following existing baseline approaches \cite{boyd2023cyborg, crum2024grains, webster2025saliency, webster2026psychophysical}. All other rows are trained using saliency-guidance with the specified saliency type, applying the established CYBORG loss formulation \cite{boyd2023cyborg} and alpha tuning scheme \cite{webster2026psychophysical}. Accuracy, Attack Presentation Classification Error Rate (APCER), and Bonafide Presentation Classification Error Rate (BPCER) are all computed at the equal error rate threshold computed over each model's respective validation set unless stated otherwise. For each metric, the highest achieved score is \textbf{bolded}. Eigenface m[0.1-0.9] saliency types describe maps produced through the Eigenface process but with their most principal components removed until the specified percent pixel-wise error compared to the original Eigenface map is reached.

\subsection{Iris PAD}
% Main results
\begin{table}[H]
\centering
\caption{Full results for models trained for iris PAD with various configurations.}
\label{tab:suppmat-iris-pad}

\definecolor{baseline}{gray}{0.90}
\definecolor{best}{RGB}{230,255,230}
\newcolumntype{C}{>{\centering\arraybackslash}X}

\begin{tabularx}{\linewidth}{l C | C C C C}
\toprule
\textbf{Saliency Type} & \textbf{Architecture} & \textbf{AUC $\uparrow$ } & \textbf{Accuracy $\uparrow$} & \textbf{APCER $\downarrow$} & \textbf{BPCER $\downarrow$}\\
\midrule

\rowcolor{baseline}
None (Baseline) & \makecell{ResNet \\ DenseNet \\ Inception} & \makecell{0.878$\pm$0.004 \\ 0.896$\pm$0.014 \\ 0.881$\pm$0.004} & \makecell{0.800$\pm$0.006 \\ 0.814$\pm$0.011 \\ 0.796$\pm$0.005} & \makecell{0.168$\pm$0.019 \\ 0.151$\pm$0.024 \\ 0.141$\pm$0.012} & \makecell{0.244$\pm$0.011 \\ \textbf{0.234$\pm$0.043} \\ 0.288$\pm$0.026} \\ \midrule
Eigenface & \makecell{ResNet \\ DenseNet \\ Inception} & \makecell{0.849$\pm$0.021 \\ 0.824$\pm$0.014 \\ 0.872$\pm$0.023} & \makecell{0.771$\pm$0.017 \\ 0.744$\pm$0.015 \\ 0.784$\pm$0.023} & \makecell{0.181$\pm$0.018 \\ 0.181$\pm$0.033 \\ 0.117$\pm$0.017} & \makecell{0.293$\pm$0.017 \\ 0.357$\pm$0.041 \\ 0.348$\pm$0.038} \\ \midrule
Fisherface & \makecell{ResNet \\ DenseNet \\ Inception} & \makecell{0.896$\pm$0.003 \\ 0.871$\pm$0.012 \\ 0.867$\pm$0.007} & \makecell{0.809$\pm$0.002 \\ 0.782$\pm$0.007 \\ 0.779$\pm$0.008} & \makecell{0.108$\pm$0.028 \\ 0.106$\pm$0.020 \\ 0.152$\pm$0.018} & \makecell{0.301$\pm$0.032 \\ 0.367$\pm$0.029 \\ 0.312$\pm$0.033} \\ \midrule
\rowcolor{best}
LDA & \makecell{ResNet \\ DenseNet \\ Inception} & \makecell{0.891$\pm$0.017 \\ \textbf{0.911$\pm$0.012} \\ 0.896$\pm$0.019} & \makecell{0.808$\pm$0.013 \\ \textbf{0.824$\pm$0.012} \\ 0.810$\pm$0.025} & \makecell{0.147$\pm$0.014 \\ \textbf{0.102$\pm$0.020} \\ 0.133$\pm$0.006} & \makecell{0.251$\pm$0.036 \\ 0.275$\pm$0.026 \\ 0.265$\pm$0.054} \\ 

\bottomrule
\end{tabularx}
\end{table}

% Ablation study
\begin{table}[H]
\centering
\caption{Full Eigenface Minus \% (M-\%) Fidelity ablation study results for models trained for iris PAD with various configurations.}
\label{tab:suppmat-iris-pad-ablation}

\definecolor{baseline}{gray}{0.90}
\definecolor{best}{RGB}{230,255,230}
\newcolumntype{C}{>{\centering\arraybackslash}X}

\begin{tabularx}{\linewidth}{l C | C C C C}
\toprule
\textbf{Saliency Type} & \textbf{Architecture} & \textbf{AUC $\uparrow$ } & \textbf{Accuracy $\uparrow$} & \textbf{APCER $\downarrow$} & \textbf{BPCER $\downarrow$}\\
\midrule

Eigenface M-0.1 & \makecell{ResNet \\ DenseNet \\ Inception} & \makecell{0.876$\pm$0.013 \\ 0.860$\pm$0.018 \\ 0.873$\pm$0.011} & \makecell{0.790$\pm$0.016 \\ 0.770$\pm$0.012 \\ 0.780$\pm$0.011} & \makecell{0.142$\pm$0.006 \\ 0.126$\pm$0.021 \\ 0.136$\pm$0.023} & \makecell{0.301$\pm$0.029 \\ 0.370$\pm$0.015 \\ 0.332$\pm$0.029} \\ \midrule
\rowcolor{best}
Eigenface M-0.2 & \makecell{ResNet \\ DenseNet \\ Inception} & \makecell{\textbf{0.887$\pm$0.008} \\ 0.870$\pm$0.034 \\ 0.875$\pm$0.011} & \makecell{\textbf{0.803$\pm$0.008} \\ 0.786$\pm$0.030 \\ 0.781$\pm$0.013} & \makecell{0.140$\pm$0.004 \\ 0.141$\pm$0.046 \\ 0.116$\pm$0.017} & \makecell{\textbf{0.273$\pm$0.022} \\ 0.313$\pm$0.021 \\ 0.356$\pm$0.033} \\ \midrule
Eigenface M-0.3 & \makecell{ResNet \\ DenseNet \\ Inception} & \makecell{0.874$\pm$0.006 \\ 0.856$\pm$0.015 \\ 0.869$\pm$0.012} & \makecell{0.785$\pm$0.002 \\ 0.765$\pm$0.016 \\ 0.781$\pm$0.012} & \makecell{0.138$\pm$0.013 \\ 0.139$\pm$0.038 \\ 0.133$\pm$0.010} & \makecell{0.317$\pm$0.021 \\ 0.364$\pm$0.018 \\ 0.334$\pm$0.015} \\ \midrule
Eigenface M-0.4 & \makecell{ResNet \\ DenseNet \\ Inception} & \makecell{0.886$\pm$0.011 \\ 0.871$\pm$0.011 \\ 0.865$\pm$0.017} & \makecell{0.802$\pm$0.014 \\ 0.785$\pm$0.012 \\ 0.773$\pm$0.013} & \makecell{0.132$\pm$0.030 \\ 0.120$\pm$0.016 \\ 0.125$\pm$0.013} & \makecell{0.287$\pm$0.040 \\ 0.340$\pm$0.027 \\ 0.363$\pm$0.041} \\ \midrule
Eigenface M-0.5 & \makecell{ResNet \\ DenseNet \\ Inception} & \makecell{0.878$\pm$0.007 \\ 0.867$\pm$0.028 \\ 0.864$\pm$0.015} & \makecell{0.790$\pm$0.004 \\ 0.776$\pm$0.024 \\ 0.779$\pm$0.014} & \makecell{0.135$\pm$0.016 \\ 0.143$\pm$0.033 \\ 0.155$\pm$0.003} & \makecell{0.311$\pm$0.011 \\ 0.332$\pm$0.034 \\ 0.309$\pm$0.030} \\ \midrule
Eigenface M-0.6 & \makecell{ResNet \\ DenseNet \\ Inception} & \makecell{0.879$\pm$0.005 \\ 0.863$\pm$0.029 \\ 0.881$\pm$0.018} & \makecell{0.789$\pm$0.005 \\ 0.779$\pm$0.026 \\ 0.782$\pm$0.014} & \makecell{0.137$\pm$0.025 \\ 0.165$\pm$0.039 \\ \textbf{0.109$\pm$0.026}} & \makecell{0.308$\pm$0.044 \\ 0.296$\pm$0.049 \\ 0.364$\pm$0.029} \\ \midrule
Eigenface M-0.7 & \makecell{ResNet \\ DenseNet \\ Inception} & \makecell{0.856$\pm$0.018 \\ 0.861$\pm$0.004 \\ 0.869$\pm$0.008} & \makecell{0.777$\pm$0.014 \\ 0.775$\pm$0.005 \\ 0.779$\pm$0.007} & \makecell{0.145$\pm$0.014 \\ 0.136$\pm$0.025 \\ 0.128$\pm$0.020} & \makecell{0.326$\pm$0.020 \\ 0.344$\pm$0.028 \\ 0.346$\pm$0.039} \\ \midrule
Eigenface M-0.8 & \makecell{ResNet \\ DenseNet \\ Inception} & \makecell{0.879$\pm$0.017 \\ 0.839$\pm$0.017 \\ 0.846$\pm$0.012} & \makecell{0.793$\pm$0.018 \\ 0.755$\pm$0.011 \\ 0.759$\pm$0.009} & \makecell{0.144$\pm$0.010 \\ 0.156$\pm$0.029 \\ 0.129$\pm$0.007} & \makecell{0.291$\pm$0.034 \\ 0.364$\pm$0.032 \\ 0.391$\pm$0.024} \\ \midrule
Eigenface M-0.9 & \makecell{ResNet \\ DenseNet \\ Inception} & \makecell{0.869$\pm$0.017 \\ 0.857$\pm$0.009 \\ 0.848$\pm$0.016} & \makecell{0.782$\pm$0.021 \\ 0.770$\pm$0.010 \\ 0.771$\pm$0.007} & \makecell{0.153$\pm$0.017 \\ 0.124$\pm$0.011 \\ 0.142$\pm$0.019} & \makecell{0.305$\pm$0.027 \\ 0.370$\pm$0.036 \\ 0.344$\pm$0.011} \\

\bottomrule
\end{tabularx}
\end{table}

\subsection{Synthetic Face Detection}
% Main results
\begin{table}[H]
\centering
\caption{Full results for models trained for synthetic face detection with various configurations.}
\label{tab:suppmat-synthface}

\definecolor{baseline}{gray}{0.90}
\definecolor{best}{RGB}{230,255,230}
\newcolumntype{C}{>{\centering\arraybackslash}X}

\begin{tabularx}{\linewidth}{l C | C C C C}
\toprule
\textbf{Saliency Type} & \textbf{Architecture} & \textbf{AUC $\uparrow$ } & \textbf{Accuracy $\uparrow$} & \textbf{APCER $\downarrow$} & \textbf{BPCER $\downarrow$}\\
\midrule

\rowcolor{baseline}
None (Baseline) & \makecell{ResNet \\ DenseNet \\ Inception} & \makecell{0.566$\pm$0.069 \\ 0.537$\pm$0.041 \\ 0.620$\pm$0.022} & \makecell{0.857$\pm$0.000 \\ 0.854$\pm$0.003 \\ 0.848$\pm$0.008} & \makecell{\textbf{0.002$\pm$0.001} \\ 0.006$\pm$0.005 \\ 0.014$\pm$0.010} & \makecell{0.990$\pm$0.006 \\ 0.984$\pm$0.008 \\ 0.983$\pm$0.006} \\ \midrule
Eigenface & \makecell{ResNet \\ DenseNet \\ Inception} & \makecell{0.510$\pm$0.006 \\ 0.426$\pm$0.046 \\ 0.598$\pm$0.050} & \makecell{0.835$\pm$0.022 \\ 0.856$\pm$0.000 \\ 0.856$\pm$0.001} & \makecell{0.035$\pm$0.034 \\ \textbf{0.002$\pm$0.001} \\ 0.005$\pm$0.002} & \makecell{\textbf{0.942$\pm$0.050} \\ 0.995$\pm$0.002 \\ 0.978$\pm$0.008} \\ \midrule
Fisherface & \makecell{ResNet \\ DenseNet \\ Inception} & \makecell{0.565$\pm$0.007 \\ 0.463$\pm$0.065 \\ \textbf{0.654$\pm$0.051}} & \makecell{0.851$\pm$0.007 \\ 0.843$\pm$0.003 \\ 0.854$\pm$0.003} & \makecell{0.013$\pm$0.012 \\ 0.022$\pm$0.004 \\ 0.008$\pm$0.004} & \makecell{0.966$\pm$0.024 \\ 0.972$\pm$0.014 \\ 0.970$\pm$0.023} \\ \midrule
\rowcolor{best}
LDA & \makecell{ResNet \\ DenseNet \\ Inception} & \makecell{0.623$\pm$0.021 \\ 0.560$\pm$0.011 \\ \textbf{0.654$\pm$0.003}} & \makecell{0.856$\pm$0.002 \\ \textbf{0.858$\pm$0.001} \\ 0.853$\pm$0.003} & \makecell{0.007$\pm$0.003 \\ 0.003$\pm$0.000 \\ 0.010$\pm$0.008} & \makecell{0.972$\pm$0.006 \\ 0.977$\pm$0.005 \\ 0.971$\pm$0.023} \\

\bottomrule
\end{tabularx}
\end{table}

\begin{table}[H]
\centering
\caption{Full Eigenface Minus \% (M-\%) Fidelity ablation study results for models trained for synthetic face detection with various configurations.}
\label{tab:suppmat-synthface-ablation}

\definecolor{baseline}{gray}{0.90}
\definecolor{best}{RGB}{230,255,230}
\newcolumntype{C}{>{\centering\arraybackslash}X}

\begin{tabularx}{\linewidth}{l C | C C C C}
\toprule
\textbf{Saliency Type} & \textbf{Architecture} & \textbf{AUC $\uparrow$ } & \textbf{Accuracy $\uparrow$} & \textbf{APCER $\downarrow$} & \textbf{BPCER $\downarrow$}\\
\midrule

Eigenface M-0.1 & \makecell{ResNet \\ DenseNet \\ Inception} & \makecell{0.480$\pm$0.027 \\ 0.407$\pm$0.012 \\ 0.617$\pm$0.051} & \makecell{0.855$\pm$0.002 \\ 0.856$\pm$0.000 \\ \textbf{0.857$\pm$0.001}} & \makecell{0.006$\pm$0.005 \\ 0.002$\pm$0.000 \\ 0.004$\pm$0.001} & \makecell{0.981$\pm$0.012 \\ 0.994$\pm$0.001 \\ 0.978$\pm$0.008} \\ \midrule
Eigenface M-0.2 & \makecell{ResNet \\ DenseNet \\ Inception} & \makecell{0.453$\pm$0.005 \\ 0.431$\pm$0.027 \\ 0.609$\pm$0.019} & \makecell{0.855$\pm$0.001 \\ 0.855$\pm$0.002 \\ 0.856$\pm$0.001} & \makecell{0.004$\pm$0.001 \\ 0.004$\pm$0.003 \\ 0.004$\pm$0.002} & \makecell{0.992$\pm$0.001 \\ 0.992$\pm$0.002 \\ 0.979$\pm$0.007} \\ \midrule
Eigenface M-0.3 & \makecell{ResNet \\ DenseNet \\ Inception} & \makecell{0.449$\pm$0.009 \\ 0.437$\pm$0.062 \\ 0.558$\pm$0.037} & \makecell{0.856$\pm$0.000 \\ 0.854$\pm$0.002 \\ \textbf{0.857$\pm$0.001}} & \makecell{0.002$\pm$0.001 \\ 0.004$\pm$0.002 \\ 0.003$\pm$0.001} & \makecell{0.993$\pm$0.002 \\ 0.993$\pm$0.000 \\ 0.986$\pm$0.002} \\ \midrule
\rowcolor{best}
Eigenface M-0.4 & \makecell{ResNet \\ DenseNet \\ Inception} & \makecell{0.441$\pm$0.016 \\ 0.428$\pm$0.028 \\ \textbf{0.627$\pm$0.020}} & \makecell{0.835$\pm$0.030 \\ 0.856$\pm$0.001 \\ 0.854$\pm$0.003} & \makecell{0.030$\pm$0.038 \\ 0.002$\pm$0.001 \\ 0.009$\pm$0.004} & \makecell{0.979$\pm$0.019 \\ 0.995$\pm$0.003 \\ \textbf{0.972$\pm$0.010}} \\ \midrule
Eigenface M-0.5 & \makecell{ResNet \\ DenseNet \\ Inception} & \makecell{0.420$\pm$0.013 \\ 0.453$\pm$0.049 \\ 0.602$\pm$0.029} & \makecell{0.856$\pm$0.001 \\ \textbf{0.857$\pm$0.000} \\ 0.854$\pm$0.003} & \makecell{0.003$\pm$0.000 \\ 0.002$\pm$0.001 \\ 0.007$\pm$0.004} & \makecell{0.995$\pm$0.002 \\ 0.991$\pm$0.003 \\ 0.984$\pm$0.003} \\ \midrule
Eigenface M-0.6 & \makecell{ResNet \\ DenseNet \\ Inception} & \makecell{0.393$\pm$0.035 \\ 0.360$\pm$0.022 \\ 0.566$\pm$0.057} & \makecell{0.847$\pm$0.010 \\ \textbf{0.857$\pm$0.000} \\ \textbf{0.857$\pm$0.000}} & \makecell{0.015$\pm$0.015 \\ \textbf{0.001$\pm$0.000} \\ 0.003$\pm$0.001} & \makecell{0.980$\pm$0.016 \\ 0.996$\pm$0.001 \\ 0.983$\pm$0.004} \\ \midrule
Eigenface M-0.7 & \makecell{ResNet \\ DenseNet \\ Inception} & \makecell{0.407$\pm$0.009 \\ 0.412$\pm$0.076 \\ 0.592$\pm$0.026} & \makecell{0.856$\pm$0.000 \\ \textbf{0.857$\pm$0.000} \\ 0.856$\pm$0.001} & \makecell{0.002$\pm$0.000 \\ \textbf{0.001$\pm$0.001} \\ 0.005$\pm$0.003} & \makecell{0.995$\pm$0.001 \\ 0.993$\pm$0.004 \\ 0.981$\pm$0.009} \\ \midrule
Eigenface M-0.8 & \makecell{ResNet \\ DenseNet \\ Inception} & \makecell{0.428$\pm$0.040 \\ 0.376$\pm$0.008 \\ 0.592$\pm$0.056} & \makecell{0.855$\pm$0.002 \\ 0.856$\pm$0.000 \\ 0.855$\pm$0.002} & \makecell{0.003$\pm$0.003 \\ 0.002$\pm$0.000 \\ 0.006$\pm$0.002} & \makecell{0.993$\pm$0.003 \\ 0.994$\pm$0.002 \\ 0.979$\pm$0.007} \\ \midrule
Eigenface M-0.9 & \makecell{ResNet \\ DenseNet \\ Inception} & \makecell{0.431$\pm$0.032 \\ 0.392$\pm$0.007 \\ 0.604$\pm$0.034} & \makecell{0.856$\pm$0.000 \\ 0.856$\pm$0.001 \\ 0.850$\pm$0.006} & \makecell{0.002$\pm$0.000 \\ 0.002$\pm$0.001 \\ 0.014$\pm$0.009} & \makecell{0.994$\pm$0.001 \\ 0.993$\pm$0.002 \\ 0.965$\pm$0.010} \\ 

\bottomrule
\end{tabularx}
\end{table}

\subsection{Fingerprint PAD}
% Main results
\begin{table}[H]
\centering
\caption{Full results for models trained for fingerprint PAD with various configurations.}
\label{tab:suppmat-fingerprint-pad}

\definecolor{baseline}{gray}{0.90}
\definecolor{best}{RGB}{230,255,230}
\newcolumntype{C}{>{\centering\arraybackslash}X}

\begin{tabularx}{\linewidth}{l C | C C C C}
\toprule
\textbf{Saliency Type} & \textbf{Architecture} & \textbf{AUC $\uparrow$ } & \textbf{Accuracy $\uparrow$} & \textbf{APCER $\downarrow$} & \textbf{BPCER $\downarrow$}\\
\midrule

\rowcolor{baseline}
None (Baseline) & \makecell{ResNet \\ DenseNet \\ Inception} & \makecell{0.930$\pm$0.003 \\ 0.918$\pm$0.016 \\ 0.946$\pm$0.007} & \makecell{0.847$\pm$0.003 \\ 0.839$\pm$0.012 \\ 0.862$\pm$0.010} & \makecell{0.240$\pm$0.014 \\ 0.275$\pm$0.025 \\ 0.231$\pm$0.014 } & \makecell{0.072$\pm$0.007 \\ 0.056$\pm$0.004 \\ 0.052$\pm$0.009} \\ \midrule
Eigenface & \makecell{ResNet \\ DenseNet \\ Inception} & \makecell{0.921$\pm$0.010 \\ 0.948$\pm$0.005 \\ 0.957$\pm$0.006} & \makecell{0.842$\pm$0.008 \\ 0.868$\pm$0.007 \\ 0.875$\pm$0.009} & \makecell{0.249$\pm$0.009 \\ 0.226$\pm$0.009 \\ 0.215$\pm$0.016} & \makecell{0.073$\pm$0.017 \\ 0.044$\pm$0.007 \\ \textbf{0.043$\pm$0.006}} \\ \midrule
Fisherface & \makecell{ResNet \\ DenseNet \\ Inception} & \makecell{0.919$\pm$0.009 \\ 0.945$\pm$0.008 \\ 0.957$\pm$0.005} & \makecell{0.835$\pm$0.008 \\ 0.862$\pm$0.017 \\ 0.877$\pm$0.001} & \makecell{0.270$\pm$0.011 \\ 0.237$\pm$0.029 \\ 0.206$\pm$0.011} & \makecell{0.067$\pm$0.019 \\ 0.046$\pm$0.007 \\ 0.046$\pm$0.012} \\ \midrule
\rowcolor{best}
LDA & \makecell{ResNet \\ DenseNet \\ Inception} & \makecell{0.930$\pm$0.003 \\ 0.926$\pm$0.005 \\ \textbf{0.962$\pm$0.002}} & \makecell{0.848$\pm$0.005 \\ 0.837$\pm$0.011 \\ \textbf{0.884$\pm$0.003}} & \makecell{0.250$\pm$0.008 \\ 0.278$\pm$0.030 \\ \textbf{0.194$\pm$0.009}} & \makecell{0.061$\pm$0.002 \\ 0.056$\pm$0.007 \\ 0.045$\pm$0.003} \\ 

\bottomrule
\end{tabularx}
\end{table}

% Ablation study
\begin{table}[H]
\centering
\caption{Full Eigenface Minus \% (M-\%) Fidelity ablation study results for models trained for fingerprint PAD with various configurations.}
\label{tab:suppmat-fingerprint-pad-ablation}

\definecolor{baseline}{gray}{0.90}
\definecolor{best}{RGB}{230,255,230}
\newcolumntype{C}{>{\centering\arraybackslash}X}

\begin{tabularx}{\linewidth}{l C | C C C C}
\toprule
\textbf{Saliency Type} & \textbf{Architecture} & \textbf{AUC $\uparrow$ } & \textbf{Accuracy $\uparrow$} & \textbf{APCER $\downarrow$} & \textbf{BPCER $\downarrow$}\\
\midrule

\rowcolor{best}
Eigenface M-0.1 & \makecell{ResNet \\ DenseNet \\ Inception} & \makecell{0.938$\pm$0.007 \\ 0.955$\pm$0.007 \\ \textbf{0.962$\pm$0.001}} & \makecell{0.865$\pm$0.007 \\ 0.884$\pm$0.010 \\ \textbf{0.885$\pm$0.003}} & \makecell{0.195$\pm$0.040 \\ \textbf{0.170$\pm$0.013} \\ 0.188$\pm$0.009} & \makecell{0.080$\pm$0.025 \\ 0.066$\pm$0.008 \\ 0.048$\pm$0.004} \\ \midrule
Eigenface M-0.2 & \makecell{ResNet \\ DenseNet \\ Inception} & \makecell{0.941$\pm$0.013 \\ 0.946$\pm$0.009 \\ 0.952$\pm$0.005} & \makecell{0.871$\pm$0.016 \\ 0.870$\pm$0.013 \\ 0.865$\pm$0.006} & \makecell{0.197$\pm$0.036 \\ 0.195$\pm$0.026 \\ 0.230$\pm$0.012} & \makecell{0.066$\pm$0.003 \\ 0.069$\pm$0.008 \\ 0.047$\pm$0.006} \\ \midrule
Eigenface M-0.3 & \makecell{ResNet \\ DenseNet \\ Inception} & \makecell{0.922$\pm$0.008 \\ 0.928$\pm$0.007 \\ 0.950$\pm$0.003} & \makecell{0.838$\pm$0.011 \\ 0.840$\pm$0.010 \\ 0.873$\pm$0.006} & \makecell{0.253$\pm$0.019 \\ 0.228$\pm$0.025 \\ 0.209$\pm$0.008} & \makecell{0.078$\pm$0.012 \\ 0.097$\pm$0.007 \\ 0.050$\pm$0.005} \\ \midrule
Eigenface M-0.4 & \makecell{ResNet \\ DenseNet \\ Inception} & \makecell{0.913$\pm$0.023 \\ 0.912$\pm$0.006 \\ 0.948$\pm$0.003} & \makecell{0.827$\pm$0.027 \\ 0.827$\pm$0.008 \\ 0.861$\pm$0.007} & \makecell{0.267$\pm$0.037 \\ 0.251$\pm$0.007 \\ 0.238$\pm$0.014} & \makecell{0.086$\pm$0.021 \\ 0.101$\pm$0.009 \\ 0.048$\pm$0.001} \\ \midrule
Eigenface M-0.5 & \makecell{ResNet \\ DenseNet \\ Inception} & \makecell{0.904$\pm$0.005 \\ 0.913$\pm$0.018 \\ 0.952$\pm$0.009} & \makecell{0.819$\pm$0.005 \\ 0.825$\pm$0.015 \\ 0.864$\pm$0.010} & \makecell{0.273$\pm$0.005 \\ 0.264$\pm$0.014 \\ 0.225$\pm$0.012} & \makecell{0.096$\pm$0.005 \\ 0.093$\pm$0.016 \\ 0.053$\pm$0.009} \\ \midrule
Eigenface M-0.6 & \makecell{ResNet \\ DenseNet \\ Inception} & \makecell{0.920$\pm$0.010 \\ 0.911$\pm$0.014 \\ 0.952$\pm$0.008} & \makecell{0.835$\pm$0.008 \\ 0.811$\pm$0.009 \\ 0.871$\pm$0.008} & \makecell{0.255$\pm$0.010 \\ 0.307$\pm$0.007 \\ 0.212$\pm$0.013} & \makecell{0.082$\pm$0.012 \\ 0.079$\pm$0.013 \\ 0.052$\pm$0.002} \\ \midrule
Eigenface M-0.7 & \makecell{ResNet \\ DenseNet \\ Inception} & \makecell{0.929$\pm$0.006 \\ 0.926$\pm$0.003 \\ 0.960$\pm$0.005} & \makecell{0.841$\pm$0.007 \\ 0.837$\pm$0.008 \\ 0.877$\pm$0.010} & \makecell{0.268$\pm$0.022 \\ 0.263$\pm$0.015 \\ 0.211$\pm$0.024} & \makecell{0.059$\pm$0.014 \\ 0.071$\pm$0.004 \\ 0.043$\pm$0.007} \\ \midrule
Eigenface M-0.8 & \makecell{ResNet \\ DenseNet \\ Inception} & \makecell{0.924$\pm$0.008 \\ 0.928$\pm$0.009 \\ 0.958$\pm$0.004} & \makecell{0.838$\pm$0.010 \\ 0.846$\pm$0.003 \\ 0.877$\pm$0.011} & \makecell{0.269$\pm$0.023 \\ 0.260$\pm$0.007 \\ 0.207$\pm$0.023} & \makecell{0.063$\pm$0.006 \\ 0.056$\pm$0.012 \\ 0.046$\pm$0.005} \\ \midrule
Eigenface M-0.9 & \makecell{ResNet \\ DenseNet \\ Inception} & \makecell{0.933$\pm$0.012 \\ 0.942$\pm$0.003 \\ 0.961$\pm$0.005} & \makecell{0.852$\pm$0.010 \\ 0.851$\pm$0.008 \\ 0.883$\pm$0.005} & \makecell{0.239$\pm$0.023 \\ 0.262$\pm$0.015 \\ 0.194$\pm$0.008} & \makecell{0.063$\pm$0.019 \\ \textbf{0.044$\pm$0.003} \\ 0.046$\pm$0.005} \\

\bottomrule
\end{tabularx}
\end{table}

\subsection{Fingerprint Vein PAD}
% Main results
\begin{table}[H]
\centering
\caption{Full results for models trained for fingerprint vein PAD with various configurations. Accuracy, APCER, and BPCER were computed at a threshold of 0.5.}
\label{tab:suppmat-fv-pad}

\definecolor{baseline}{gray}{0.90}
\definecolor{best}{RGB}{230,255,230}
\newcolumntype{C}{>{\centering\arraybackslash}X}

\begin{tabularx}{\linewidth}{l C | C C C C}
\toprule
\textbf{Saliency Type} & \textbf{Architecture} & \textbf{AUC $\uparrow$ } & \textbf{Accuracy $\uparrow$} & \textbf{APCER $\downarrow$} & \textbf{BPCER $\downarrow$}\\
\midrule

\rowcolor{baseline}
None (Baseline) & \makecell{ResNet \\ DenseNet \\ Inception} & \makecell{\textbf{1.000$\pm$0.000} \\ \textbf{1.000$\pm$0.000} \\ \textbf{1.000$\pm$0.000}} & \makecell{0.921$\pm$0.011 \\ 0.927$\pm$0.044 \\ 0.926$\pm$0.017} & \makecell{0.107$\pm$0.015 \\ 0.099$\pm$0.060 \\ 0.101$\pm$0.023} & \makecell{\textbf{0.000$\pm$0.000} \\ \textbf{0.000$\pm$0.000} \\ \textbf{0.000$\pm$0.000}} \\ \midrule
Eigenface & \makecell{ResNet \\ DenseNet \\ Inception} & \makecell{\textbf{1.000$\pm$0.000} \\ \textbf{1.000$\pm$0.000} \\ \textbf{1.000$\pm$0.000}} & \makecell{0.921$\pm$0.010 \\ 0.904$\pm$0.008 \\ 0.920$\pm$0.013} & \makecell{0.107$\pm$0.013 \\ 0.130$\pm$0.010 \\ 0.109$\pm$0.018} & \makecell{\textbf{0.000$\pm$0.000} \\ \textbf{0.000$\pm$0.000} \\ \textbf{0.000$\pm$0.000}} \\ \midrule
Fisherface & \makecell{ResNet \\ DenseNet \\ Inception} & \makecell{\textbf{1.000$\pm$0.000} \\ \textbf{1.000$\pm$0.000} \\ \textbf{1.000$\pm$0.000}} & \makecell{0.909$\pm$0.006 \\ 0.927$\pm$0.018 \\ 0.930$\pm$0.033} & \makecell{0.124$\pm$0.008 \\ 0.099$\pm$0.024 \\ 0.096$\pm$0.045} & \makecell{\textbf{0.000$\pm$0.000} \\ \textbf{0.000$\pm$0.000} \\ \textbf{0.000$\pm$0.000}} \\ \midrule
\rowcolor{best}
LDA & \makecell{ResNet \\ DenseNet \\ Inception} & \makecell{\textbf{1.000$\pm$0.000} \\ \textbf{1.000$\pm$0.000} \\ \textbf{1.000$\pm$0.000}} & \makecell{0.942$\pm$0.008 \\ 0.887$\pm$0.039 \\ \textbf{0.998$\pm$0.002}} & \makecell{0.078$\pm$0.010 \\ 0.154$\pm$0.053 \\ \textbf{0.002$\pm$0.002}} & \makecell{0.003$\pm$0.000 \\ \textbf{0.000$\pm$0.000} \\ 0.003$\pm$0.002} \\

\bottomrule
\end{tabularx}
\end{table}

% Ablation study
\begin{table}[H]
\centering
\caption{Full Eigenface Minus \% (M-\%) Fidelity ablation study results for models trained for fingerprint vein PAD with various configurations. Accuracy, APCER, and BPCER were computed at a threshold of 0.5.}
\label{tab:suppmat-fv-pad-ablation}

\definecolor{baseline}{gray}{0.90}
\definecolor{best}{RGB}{230,255,230}
\newcolumntype{C}{>{\centering\arraybackslash}X}

\begin{tabularx}{\linewidth}{l C | C C C C}
\toprule
\textbf{Saliency Type} & \textbf{Architecture} & \textbf{AUC $\uparrow$ } & \textbf{Accuracy $\uparrow$} & \textbf{APCER $\downarrow$} & \textbf{BPCER $\downarrow$}\\
\midrule

\rowcolor{best}
Eigenface M-0.1 & \makecell{ResNet \\ DenseNet \\ Inception} & \makecell{\textbf{1.000$\pm$0.000} \\ \textbf{1.000$\pm$0.000} \\ \textbf{1.000$\pm$0.000}} & \makecell{0.907$\pm$0.015 \\ 0.910$\pm$0.005 \\ \textbf{0.999$\pm$0.001}} & \makecell{0.126$\pm$0.021 \\ 0.123$\pm$0.006 \\ \textbf{0.000$\pm$0.000}} & \makecell{\textbf{0.000$\pm$0.000} \\ \textbf{0.000$\pm$0.000} \\ 0.006$\pm$0.004} \\ \midrule
Eigenface M-0.2 & \makecell{ResNet \\ DenseNet \\ Inception} & \makecell{\textbf{1.000$\pm$0.000} \\ \textbf{1.000$\pm$0.000} \\ \textbf{1.000$\pm$0.000}} & \makecell{0.913$\pm$0.006 \\ 0.914$\pm$0.007 \\ 0.993$\pm$0.003} & \makecell{0.118$\pm$0.009 \\ 0.118$\pm$0.009 \\ 0.009$\pm$0.004} & \makecell{\textbf{0.000$\pm$0.000} \\ \textbf{0.000$\pm$0.000} \\ 0.001$\pm$0.001} \\ \midrule
Eigenface M-0.3 & \makecell{ResNet \\ DenseNet \\ Inception} & \makecell{\textbf{1.000$\pm$0.000} \\ \textbf{1.000$\pm$0.000} \\ \textbf{1.000$\pm$0.000}} & \makecell{0.896$\pm$0.008 \\ 0.915$\pm$0.018 \\ 0.995$\pm$0.005} & \makecell{0.141$\pm$0.010 \\ 0.116$\pm$0.024 \\ 0.007$\pm$0.007} & \makecell{\textbf{0.000$\pm$0.000} \\ \textbf{0.000$\pm$0.000} \\ 0.001$\pm$0.001} \\ \midrule
Eigenface M-0.4 & \makecell{ResNet \\ DenseNet \\ Inception} & \makecell{\textbf{1.000$\pm$0.000} \\ \textbf{1.000$\pm$0.000} \\ \textbf{1.000$\pm$0.000}} & \makecell{0.927$\pm$0.020 \\ 0.936$\pm$0.028 \\ 0.984$\pm$0.014} & \makecell{0.100$\pm$0.027 \\ 0.088$\pm$0.038 \\ 0.022$\pm$0.019} & \makecell{\textbf{0.000$\pm$0.000} \\ \textbf{0.000$\pm$0.000} \\ 0.002$\pm$0.003} \\ \midrule
Eigenface M-0.5 & \makecell{ResNet \\ DenseNet \\ Inception} & \makecell{\textbf{1.000$\pm$0.000} \\ \textbf{1.000$\pm$0.000} \\ \textbf{1.000$\pm$0.000}} & \makecell{0.969$\pm$0.022 \\ 0.919$\pm$0.016 \\ 0.971$\pm$0.019} & \makecell{0.042$\pm$0.030 \\ 0.111$\pm$0.022 \\ 0.039$\pm$0.026} & \makecell{\textbf{0.000$\pm$0.000} \\ \textbf{0.000$\pm$0.000} \\ 0.002$\pm$0.003} \\ \midrule
Eigenface M-0.6 & \makecell{ResNet \\ DenseNet \\ Inception} & \makecell{\textbf{1.000$\pm$0.000} \\ \textbf{1.000$\pm$0.000} \\ \textbf{1.000$\pm$0.000}} & \makecell{0.929$\pm$0.017 \\ 0.918$\pm$0.029 \\ 0.978$\pm$0.016} & \makecell{0.097$\pm$0.023 \\ 0.112$\pm$0.040 \\ 0.030$\pm$0.022} & \makecell{\textbf{0.000$\pm$0.000} \\ \textbf{0.000$\pm$0.000} \\ \textbf{0.000$\pm$0.000}} \\ \midrule
Eigenface M-0.7 & \makecell{ResNet \\ DenseNet \\ Inception} & \makecell{\textbf{1.000$\pm$0.000} \\ \textbf{1.000$\pm$0.000} \\ \textbf{1.000$\pm$0.000}} & \makecell{0.930$\pm$0.013 \\ 0.937$\pm$0.030 \\ 0.986$\pm$0.010} & \makecell{0.095$\pm$0.017 \\ 0.086$\pm$0.041 \\ 0.017$\pm$0.015} & \makecell{\textbf{0.000$\pm$0.000} \\ \textbf{0.000$\pm$0.000} \\ 0.006$\pm$0.006} \\ \midrule
Eigenface M-0.8 & \makecell{ResNet \\ DenseNet \\ Inception} & \makecell{\textbf{1.000$\pm$0.000} \\ \textbf{1.000$\pm$0.000} \\ \textbf{1.000$\pm$0.000}} & \makecell{0.935$\pm$0.004 \\ 0.925$\pm$0.009 \\ 0.983$\pm$0.021} & \makecell{0.089$\pm$0.005 \\ 0.102$\pm$0.013 \\ 0.022$\pm$0.029} & \makecell{\textbf{0.000$\pm$0.000} \\ \textbf{0.000$\pm$0.000} \\ 0.004$\pm$0.001} \\ \midrule
Eigenface M-0.9 & \makecell{ResNet \\ DenseNet \\ Inception} & \makecell{\textbf{1.000$\pm$0.000} \\ \textbf{1.000$\pm$0.000} \\ \textbf{1.000$\pm$0.000}} & \makecell{0.936$\pm$0.003 \\ 0.914$\pm$0.010 \\ 0.969$\pm$0.018} & \makecell{0.086$\pm$0.003 \\ 0.118$\pm$0.013 \\ 0.042$\pm$0.025} & \makecell{0.002$\pm$0.003 \\ \textbf{0.000$\pm$0.000} \\ 0.004$\pm$0.001} \\

\bottomrule
\end{tabularx}
\end{table}

\subsection{Identification Document PAD}
% Main results
\begin{table}[H]
\centering
\caption{Full results for models trained for identification document PAD with various configurations.}
\label{tab:suppmat-id-pad}

\definecolor{baseline}{gray}{0.90}
\definecolor{best}{RGB}{230,255,230}
\newcolumntype{C}{>{\centering\arraybackslash}X}

\begin{tabularx}{\linewidth}{l C | C C C C}
\toprule
\textbf{Saliency Type} & \textbf{Architecture} & \textbf{AUC $\uparrow$ } & \textbf{Accuracy $\uparrow$} & \textbf{APCER $\downarrow$} & \textbf{BPCER $\downarrow$}\\
\midrule

\rowcolor{baseline}
None (Baseline) & \makecell{ResNet \\ DenseNet \\ Inception} & \makecell{0.566$\pm$0.080 \\ 0.804$\pm$0.229 \\ 0.884$\pm$0.072} & \makecell{0.565$\pm$0.065 \\ 0.779$\pm$0.207 \\ 0.812$\pm$0.055} & \makecell{0.438$\pm$0.068 \\ 0.218$\pm$0.189 \\ 0.210$\pm$0.065} & \makecell{0.430$\pm$0.066 \\ 0.226$\pm$0.233 \\ 0.157$\pm$0.054} \\ \midrule
Eigenface & \makecell{ResNet \\ DenseNet \\ Inception} & \makecell{0.806$\pm$0.051 \\ 0.929$\pm$0.015 \\ 0.887$\pm$0.027} & \makecell{0.732$\pm$0.045 \\ 0.842$\pm$0.013 \\ 0.796$\pm$0.018} & \makecell{0.282$\pm$0.038 \\ 0.172$\pm$0.011 \\ 0.239$\pm$0.012} & \makecell{0.248$\pm$0.057 \\ 0.138$\pm$0.028 \\ 0.155$\pm$0.046}  \\ \midrule
Fisherface & \makecell{ResNet \\ DenseNet \\ Inception} & \makecell{0.870$\pm$0.013 \\ 0.940$\pm$0.030 \\ 0.908$\pm$0.024} & \makecell{0.803$\pm$0.019 \\ 0.862$\pm$0.035 \\ 0.821$\pm$0.030} & \makecell{0.226$\pm$0.022 \\ 0.173$\pm$0.032 \\ 0.218$\pm$0.039} & \makecell{0.159$\pm$0.014 \\ \textbf{0.091$\pm$0.039} \\ 0.126$\pm$0.017}  \\ \midrule
\rowcolor{best}
LDA & \makecell{ResNet \\ DenseNet \\ Inception} & \makecell{0.871$\pm$0.056 \\ \textbf{0.950$\pm$0.014} \\ 0.893$\pm$0.044} & \makecell{0.777$\pm$0.060 \\ \textbf{0.873$\pm$0.023} \\ 0.799$\pm$0.055} & \makecell{0.256$\pm$0.056 \\ \textbf{0.148$\pm$0.035} \\ 0.230$\pm$0.055} & \makecell{0.178$\pm$0.065 \\ 0.099$\pm$0.011 \\ 0.162$\pm$0.055} \\

\bottomrule
\end{tabularx}
\end{table}

% Ablation study
\begin{table}[H]
\centering
\caption{Full Eigenface Minus \% (M-\%) Fidelity ablation study results for models trained for identification document PAD with various configurations.}
\label{tab:suppmat-id-pad-ablation}

\definecolor{baseline}{gray}{0.90}
\definecolor{best}{RGB}{230,255,230}
\newcolumntype{C}{>{\centering\arraybackslash}X}

\begin{tabularx}{\linewidth}{l C | C C C C}
\toprule
\textbf{Saliency Type} & \textbf{Architecture} & \textbf{AUC $\uparrow$ } & \textbf{Accuracy $\uparrow$} & \textbf{APCER $\downarrow$} & \textbf{BPCER $\downarrow$}\\
\midrule

Eigenface M-0.1 & \makecell{ResNet \\ DenseNet \\ Inception} & \makecell{0.821$\pm$0.031 \\ 0.886$\pm$0.017 \\ 0.868$\pm$0.034} & \makecell{0.747$\pm$0.020 \\ 0.795$\pm$0.022 \\ 0.773$\pm$0.029} & \makecell{0.264$\pm$0.019 \\ 0.231$\pm$0.024 \\ 0.264$\pm$0.024} & \makecell{0.238$\pm$0.025 \\ 0.169$\pm$0.020 \\ 0.178$\pm$0.037} \\ \midrule
Eigenface M-0.2 & \makecell{ResNet \\ DenseNet \\ Inception} & \makecell{0.845$\pm$0.046 \\ 0.891$\pm$0.033 \\ 0.855$\pm$0.006} & \makecell{0.767$\pm$0.041 \\ 0.808$\pm$0.040 \\ 0.771$\pm$0.008} & \makecell{0.257$\pm$0.034 \\ 0.202$\pm$0.042 \\ 0.258$\pm$0.013} & \makecell{0.201$\pm$0.052 \\ 0.179$\pm$0.051 \\ 0.189$\pm$0.009} \\ \midrule
Eigenface M-0.3 & \makecell{ResNet \\ DenseNet \\ Inception} & \makecell{0.818$\pm$0.028 \\ 0.875$\pm$0.027 \\ 0.794$\pm$0.017} & \makecell{0.747$\pm$0.025 \\ 0.794$\pm$0.025 \\ 0.718$\pm$0.025} & \makecell{0.267$\pm$0.016 \\ 0.221$\pm$0.035 \\ 0.292$\pm$0.032} & \makecell{0.234$\pm$0.040 \\ 0.186$\pm$0.035 \\ 0.268$\pm$0.015} \\ \midrule
\rowcolor{best}
Eigenface M-0.4 & \makecell{ResNet \\ DenseNet \\ Inception} & \makecell{0.804$\pm$0.001 \\ \textbf{0.923$\pm$0.016} \\ 0.790$\pm$0.017} & \makecell{0.728$\pm$0.013 \\ \textbf{0.845$\pm$0.020} \\ 0.707$\pm$0.015} & \makecell{0.288$\pm$0.009 \\ \textbf{0.166$\pm$0.026} \\ 0.321$\pm$0.030} & \makecell{0.252$\pm$0.027 \\ \textbf{0.140$\pm$0.036} \\ 0.256$\pm$0.017} \\ \midrule
Eigenface M-0.5 & \makecell{ResNet \\ DenseNet \\ Inception} & \makecell{0.760$\pm$0.045 \\ 0.840$\pm$0.019 \\ 0.740$\pm$0.038} & \makecell{0.689$\pm$0.033 \\ 0.755$\pm$0.022 \\ 0.681$\pm$0.016} & \makecell{0.328$\pm$0.029 \\ 0.262$\pm$0.034 \\ 0.335$\pm$0.024} & \makecell{0.288$\pm$0.043 \\ 0.222$\pm$0.010 \\ 0.298$\pm$0.039} \\ \midrule
Eigenface M-0.6 & \makecell{ResNet \\ DenseNet \\ Inception} & \makecell{0.757$\pm$0.055 \\ 0.839$\pm$0.046 \\ 0.614$\pm$0.037} & \makecell{0.697$\pm$0.043 \\ 0.748$\pm$0.046 \\ 0.582$\pm$0.018} & \makecell{0.320$\pm$0.033 \\ 0.284$\pm$0.056 \\ 0.460$\pm$0.026} & \makecell{0.281$\pm$0.056 \\ 0.210$\pm$0.036 \\ 0.363$\pm$0.035} \\ \midrule
Eigenface M-0.7 & \makecell{ResNet \\ DenseNet \\ Inception} & \makecell{0.831$\pm$0.013 \\ 0.891$\pm$0.022 \\ 0.736$\pm$0.035} & \makecell{0.749$\pm$0.007 \\ 0.806$\pm$0.037 \\ 0.682$\pm$0.030} & \makecell{0.258$\pm$0.017 \\ 0.215$\pm$0.052 \\ 0.331$\pm$0.021} & \makecell{0.241$\pm$0.009 \\ 0.167$\pm$0.018 \\ 0.300$\pm$0.047} \\ \midrule
Eigenface M-0.8 & \makecell{ResNet \\ DenseNet \\ Inception} & \makecell{0.810$\pm$0.018 \\ 0.864$\pm$0.020 \\ 0.824$\pm$0.012} & \makecell{0.738$\pm$0.009 \\ 0.777$\pm$0.019 \\ 0.748$\pm$0.004} & \makecell{0.282$\pm$0.005 \\ 0.236$\pm$0.026 \\ 0.281$\pm$0.010} & \makecell{0.234$\pm$0.015 \\ 0.205$\pm$0.025 \\ 0.214$\pm$0.015} \\ \midrule
Eigenface M-0.9 & \makecell{ResNet \\ DenseNet \\ Inception} & \makecell{0.809$\pm$0.015 \\ 0.897$\pm$0.028 \\ 0.821$\pm$0.033} & \makecell{0.727$\pm$0.022 \\ 0.802$\pm$0.027 \\ 0.739$\pm$0.029} & \makecell{0.284$\pm$0.028 \\ 0.229$\pm$0.026 \\ 0.300$\pm$0.017} & \makecell{0.257$\pm$0.015 \\ 0.156$\pm$0.029 \\ 0.208$\pm$0.046} \\

\bottomrule
\end{tabularx}
\end{table}

\end{document}